\documentclass[12pt]{article}

\pdfoutput=1

\usepackage[utf8]{inputenc}
\usepackage{amsmath,amssymb}
\usepackage{graphicx}
\usepackage{hyperref}
\usepackage{cite}
\usepackage{geometry}
\usepackage{float}
\usepackage{authblk} 

\usepackage[font=footnotesize,labelfont=bf]{caption}

\title{\textbf{Embodied World Models Emerge from Navigational Task in Open-Ended Environments}}

\author[1]{Li Jin}
\author[2]{Liu Jia}

\affil[1]{Tsinghua Laboratory of Brain and Intelligence \\
           \texttt{electrixoul@outlook.com}}
\affil[2]{Tsinghua Laboratory of Brain and Intelligence \\
           \texttt{liujiaTHU@tsinghua.edu.cn}}

\date{} 

\begin{document}

\maketitle

\section*{Abstract}

Spatial reasoning in partially observable environments has often been approached through passive predictive models, yet theories of embodied cognition suggest that genuinely useful representations arise only when perception is tightly coupled to action. Here we ask whether a recurrent agent, trained solely by sparse rewards to solve procedurally generated planar mazes, can autonomously internalize metric concepts such as direction, distance and obstacle layout. After training, the agent consistently produces near-optimal paths in unseen mazes, behavior that hints at an underlying spatial model.  

To probe this possibility, we cast the closed agent–environment loop as a hybrid dynamical system, identify stable limit cycles in its state space, and characterize behavior with a Ridge Representation that embeds whole trajectories into a common metric space. Canonical correlation analysis exposes a robust linear alignment between neural and behavioral manifolds, while targeted perturbations of the most informative neural dimensions sharply degrade navigation performance.  

Taken together, these dynamical, representational, and causal signatures show that sustained sensorimotor interaction is sufficient for the spontaneous emergence of compact, embodied world models, providing a principled path toward interpretable and transferable navigation policies.

\medskip
\hrule
\medskip

\section{Introduction}

Intelligent behaviour hinges on internal \emph{world model}—a representation that predicts how the environment will change in response to the agent’s own actions.  Early cognitive–computational frameworks attempted to build such models from \emph{passive} observation, separating perception from control \cite{Friston2016,Friston2021,Wiedermann2006}.  By contrast, the theory of \emph{embodied cognition} argues that durable abstractions arise only when perception and action are inseparable: spatial notions such as “\textit{up}’’ and “\textit{down}’’ do not emerge from merely viewing elevator arrows, but from repeated bodily encounters with gravity and the effort of moving against it \cite{Barsalou2008,Glenberg2002}.  Although modern reinforcement-learning systems can solve intricate navigation tasks, it remains contentious whether their hidden states encode metric structure or merely cache brittle stimulus–response associations.

\begin{figure}[H]
    \centering
    \includegraphics[width=1.0\textwidth]{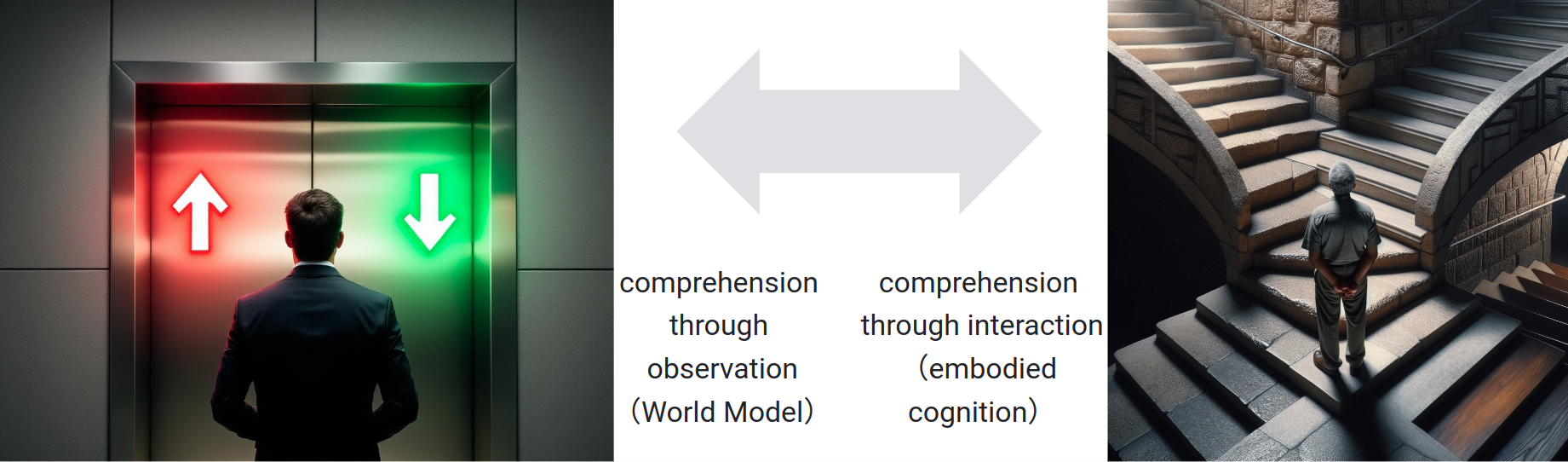}
    \caption{Conceptual illustration contrasting comprehension through observation (World Model) with comprehension through interaction (Embodied Cognition).
 On the left, the elevator with upward/downward indicators and a figure facing away symbolizes the “observation-based” paradigm, where an agent relies primarily on external cues to construct a world model. On the right, the figure on a spiraling staircase underscores the importance of physical actions and sensorimotor feedback—hallmarks of embodied cognition. The bidirectional arrow in the center highlights the transition between these two frameworks, suggesting how direct interaction can yield more deeply grounded internal representations in complex environments.}
    \label{fig:my_figure}
\end{figure}

\bigskip
\noindent
\textbf{A key question arises:} \emph{when an agent actively influences its surroundings while perceiving only a local view, does it automatically develop detailed, stable representations of direction, obstacles and path differences?}  An affirmative answer would imply genuine \emph{internalisation} of spatial knowledge; a negative one would suggest that competent performance can still rest on superficial heuristics that may fail outside the training distribution.

\bigskip
\noindent
We address this question by training a gated-recurrent agent with a meta-reinforcement-learning outer loop \cite{Duan2017,Wang2018} on thousands of procedurally generated $10 \times 10$ mazes released by an open-ended curriculum generator \cite{OpenEnded2021}.  Sparse rewards and a $3 \times 3$ field of view force the network to accumulate evidence over time and to reuse its hidden state after each goal reset, mirroring the incremental learning posited by embodied-cognition theories.  Across 10\,000 unseen layouts the agent achieves a $96\%$ reach rate and a final-trial path optimality of $0.92$, indicating that rote memorisation alone cannot explain its behaviour.

\bigskip
\noindent
To uncover what the network has learned, we (i) model the closed agent–environment loop as a \emph{Hybrid Dynamical System}, locating limit cycles whose negative Lyapunov exponents mark attractor-level strategies; (ii) embed entire trajectories with a \emph{Ridge Representation} that places behaviour in a metric space; and (iii) apply \emph{Canonical Correlation Analysis} to align this space with high-dimensional hidden states.  The first ten canonical modes each exceed $0.8$ correlation, and selectively perturbing those neural dimensions doubles path length and halves success, demonstrating causal necessity.  These dynamical, representational and causal results show that sustained sensorimotor interaction alone can drive a recurrent network to construct compact, interpretable embodied world models, offering both mechanistic insight into spatial concept formation and a practical diagnostic toolkit for future navigation agents.

\section{Results}

Having outlined the research motivation and overall framework in the Introduction, we now present the study’s core experimental findings. Detailed information on training procedures, navigation environment design, and feature extraction for both neural and behavioral data is provided in the \textbf{Methods} section. Here, we focus on the agent’s behavior in random maze tasks and the strong coupling we observe between neural and behavioral states.

\subsection{Emergence of Deep Behavioral Understanding}

This section explores the extent to which agents can develop deeper understanding in planar maze tasks, viewed from both navigation performance and neural representations. In \textbf{Section~2.1.1}, we describe the agent’s high-level performance in random mazes and explain why external behavior alone does not confirm genuine mastery of spatial properties such as direction and distance. In \textbf{Section~2.1.2}, we introduce our \textbf{Hybrid Dynamical Systems (HDS)} approach for gathering more stable and controlled neural data, avoiding biases from suboptimal or random trajectories. Finally, \textbf{Section~2.1.3} presents PCA-based visualizations of the HDS-sampled neural states, revealing ring-like structures suggestive of spatial encoding. These findings set the stage for further neural-behavioral comparisons.

\subsubsection{Overall High-Level Performance}

\paragraph{Efficient Navigation in Random Mazes}

After multiple generations (often thousands to tens of thousands) of evolution-reinforcement hybrid training, agents showed marked improvements in navigating $10 \times 10$ randomly generated mazes. As detailed in \emph{Methods}, around 8{,}000 generations were sufficient to reduce the average path length significantly and boost success rates, allowing agents to consistently find near-optimal routes in a variety of configurations. In comparisons with baselines that excluded meta-reinforcement learning or recurrent architectures, our approach achieved faster convergence and better final performance.

\begin{figure}[H]
    \centering
    \includegraphics[width=1.0\textwidth]{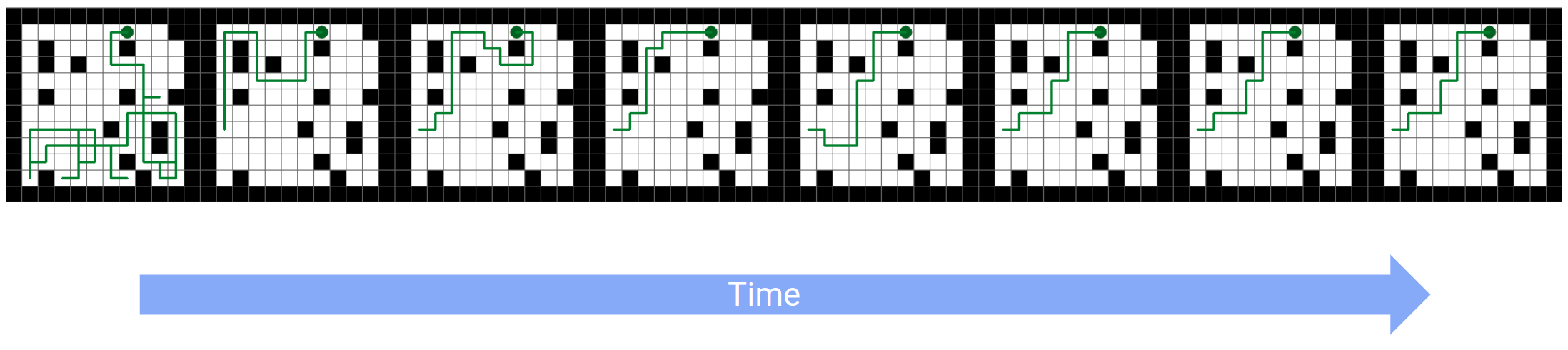}
    \caption{Illustration of efficient navigation in random mazes over time.
 Each panel displays a distinct 10×10 maze with black squares indicating obstacles and white squares denoting traversable cells. The agent’s path (green line) begins at the green circle (start) and evolves across consecutive attempts, progressively shortening and refining as the agent internalizes key layout information. By preserving its recurrent hidden state after each goal reach, the agent rapidly transitions from exploratory, sometimes circuitous routes to near-optimal navigation strategies. The horizontal axis depicts the temporal sequence of attempts, highlighting how path length and detours decrease with continued interactions.}
    \label{fig:my_figure}
\end{figure}

Nevertheless, producing relatively short paths in most random mazes and maintaining high success rates does not automatically signify “deep understanding” of key elements like direction, distance, or obstacle arrangements. Observing agents quickly reach goals over several trials does not clarify whether they are relying on memorized patterns or truly encoding spatial relationships. Consequently, more nuanced neural-behavioral analyses are required to uncover the internal processes that support these navigation outcomes.

\paragraph{Repeated Trials \& Partial Reset}

Our environment incorporates a \textbf{Goal-Reset mechanism}: whenever an agent reaches the goal, only its physical position is reset to the start, leaving the GRU hidden state \emph{unchanged}. Empirical observations show that this design greatly accelerates learning in the same maze: after initially taking longer or circuitous paths, agents rapidly refine and converge on shorter routes, often uncovering shortcuts around obstacles. Random networks or models without gating mechanisms rarely exhibit this pronounced “post-exploration path shortening,” indicating that recurrent units and meta-reinforcement learning enable agents to \emph{internalize} essential environmental information.

Still, even when agents perform efficiently, frequently reuse similar routes, and display consistent external actions, we cannot conclusively state that the network encodes geometric concepts such as “direction” and “distance.” To address potential biases from chance or suboptimal trajectories, we next describe how we use \textbf{Hybrid Dynamical Systems (HDS)} to collect neural states corresponding to near-optimal strategies, laying a stronger basis for subsequent visualization and quantitative analysis.

\subsubsection{Sampling Stable Neural Representations (via HDS)}

\paragraph{Why Not Direct Environment Trajectories}

A standard method might be to record neural activations directly from agent trajectories in the environment. However, for embodied tasks, this approach faces several obstacles:

\begin{enumerate}
    \item \textbf{Suboptimal or random paths}: During training or exploration, agents do not always follow the best routes, making it difficult to capture all critical strategy elements.
    \item \textbf{Incomplete or noisy data}: Variations in environment settings, obstacle placement, and exploration behaviors can prevent gathering the specific paths needed for thorough analysis.
    \item \textbf{Closed-loop causality}: Merely observing that “the agent traversed a certain route” does not clarify whether this was the network’s deliberate choice or simply a passive response to environmental cues.
\end{enumerate}

To overcome these challenges, we adopt the \textbf{Hybrid Dynamical Systems (HDS)} framework, which unifies discrete maze coordinates and the network’s hidden states (see \emph{Methods}). We then use cyclic stimulation and related techniques to guide agents into optimal strategy loops. This approach enables us to acquire \emph{steady-state} or \emph{near-optimal} neural activations in a controlled manner, minimizing transient or noisy components. As a result, we obtain a more reliable foundation for analyzing how neural states correspond to behavior.

\begin{figure}[H]
    \centering
    \includegraphics[width=1.0\textwidth]{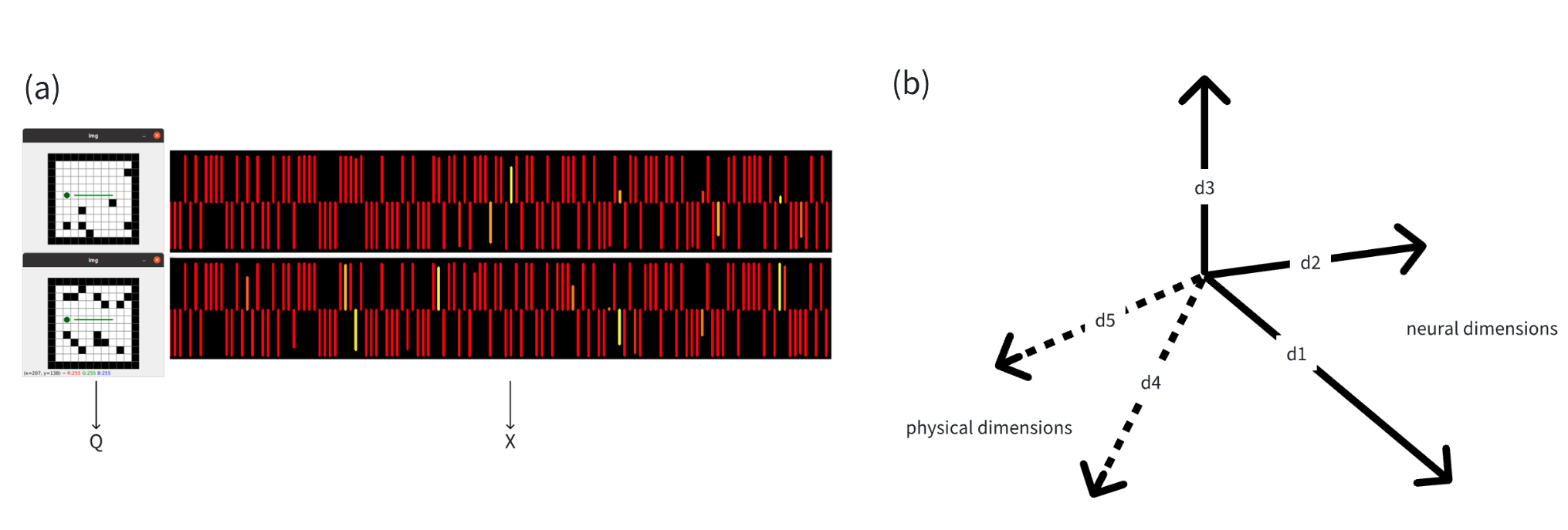}
    \caption{Rationale for hybrid dynamical sampling beyond direct environment trajectories.
(a) Illustration of raw maze trajectories (Q) alongside corresponding neural activations (X) for two different maze layouts. The maze panels (left) show agent movement (green circles and lines) in grids with obstacles (black squares); at each step, a locally observed state triggers updates in the network’s hidden units, depicted here as vertical bars (red/yellow) reflecting activation magnitudes. Directly recording such environment trajectories can introduce biases from suboptimal or exploratory paths, making it difficult to isolate the agent’s intended strategy.
(b) Conceptual diagram of the joint (Q×X) space, in which both physical coordinates (d4,d5) and neural dimensions (d1,d2,d3) jointly evolve in a closed loop. Examining this higher-dimensional hybrid system enables cleaner identification of stable or near-optimal strategy attractors, overcoming limitations of naive trajectory sampling.}
    \label{fig:my_figure}
\end{figure}

\paragraph{Initial Observations}

Within the HDS framework, we performed cyclic stimulation on well-trained agents by repeatedly feeding them fixed observation sequences, or we orchestrated multiple back-and-forth traversals in a maze to identify potential limit cycles (see \emph{Methods} for “Cyclic Stimulation and Lyapunov Exponents”). We observed that as agents became familiar with specific layouts, they often entered stable closed-loop cycles in which both the environment and network states evolved in repeating patterns. Compared to data gathered under ordinary exploration, these \textbf{HDS-sampled} neural states better reflect the agent’s “deep internalization” of an optimal action cycle and are less influenced by random or exploratory effects.

\paragraph{}
Here, we only summarize why HDS sampling offers a cleaner view of neural activations associated with optimal strategies; the \textbf{Methods} section provides extensive details on limit cycle detection, stability metrics, cyclic stimulation configurations, and initial states. Crucially, these stable neural activations form the primary dataset for subsequent visualization and high-dimensional comparisons. If lower-dimensional projections of these states reveal distinct patterns for direction or other spatial factors, that strongly implies an “embodied understanding” embedded in the agent’s network.

\begin{figure}[H]
    \centering
    \includegraphics[width=1.0\textwidth]{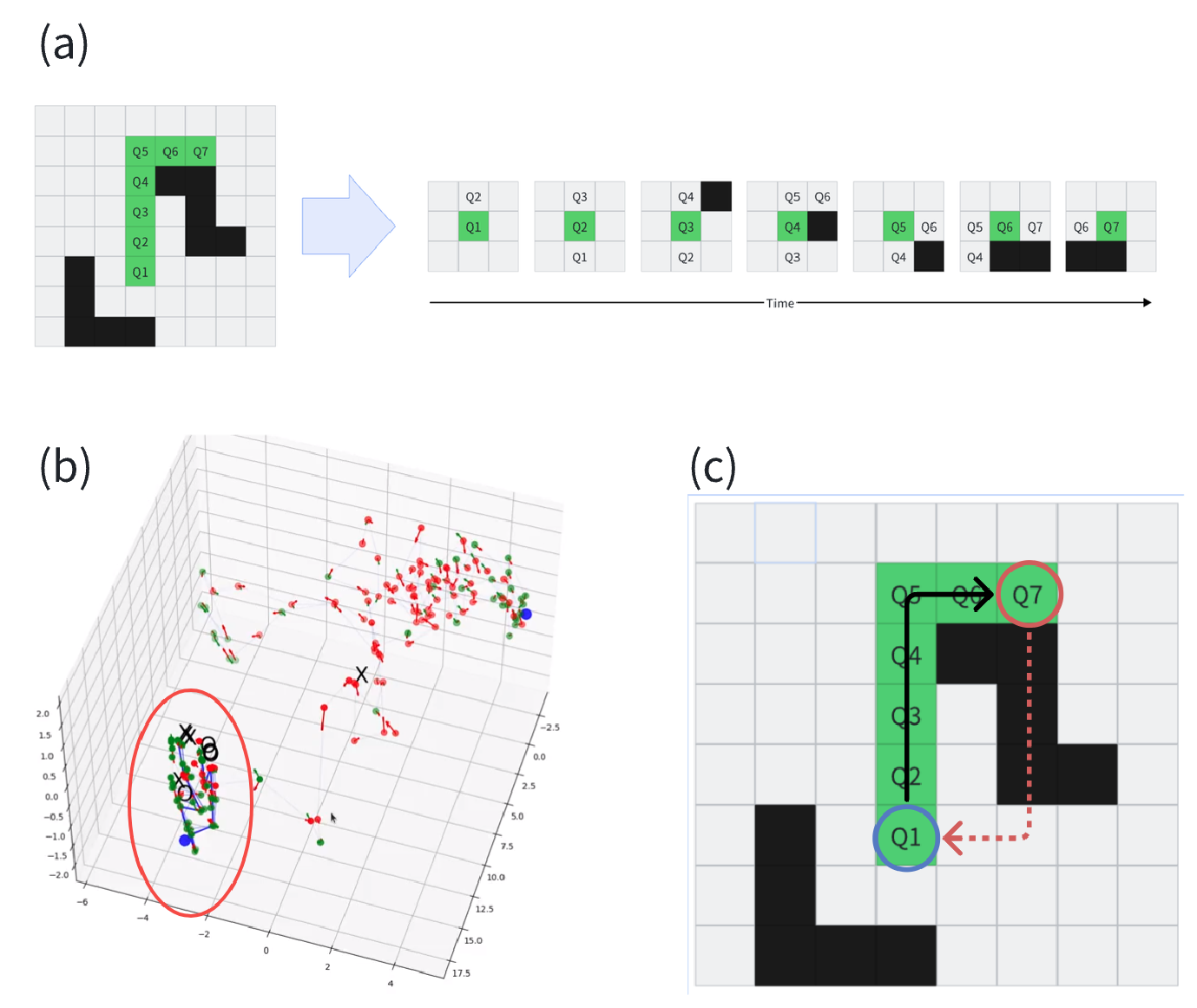}
    \caption{Initial observations of limit-cycle sampling in the hybrid Q×X space.
(a) Schematic of a partially obstructed 10×10 maze, illustrating sequential partial resets. The agent progresses from Q1 to Q7 (green cells) while retaining its hidden state between attempts, thereby refining its path.
(b) Sample 3D projection (e.g., via PCA) of hidden-state vectors recorded during these traversals. Clusters or recurrent loops in the activation space (one highlighted by the red ellipse) suggest an emerging limit cycle as the agent stabilizes its route.
(c) Maze illustration showing how the agent—upon reaching Q7—is reset back to Q1 while preserving its internal states. This cyclical “Q1→Q7→Q1” process creates a closed-loop trajectory in physical space, reinforcing a stable limit cycle both in the environment and the agent’s neural representation.}
    \label{fig:my_figure}
\end{figure}

\subsubsection{PCA of HDS-Sampled Neural States}

\paragraph{Low-Dimensional Ring Structures}

After normalizing the HDS-sampled hidden states (commonly 128-dimensional GRU vectors) by time step, we applied Principal Component Analysis (PCA) to project them into 2D or 3D for visualization. In many cases, these activations showed \emph{ring-shaped} or \emph{smoothly curved} clusters around a central region. Color-coding points by “direction” or “distance to goal” often produced polar-coordinate-like gradients: points corresponding to the same direction lay close together, while different directions occupied distinct arcs on the ring. Proximity to the goal sometimes mapped onto inner or outer bands of the ring. Such patterns suggest that the agent has developed a distributed representation of core spatial variables, including direction and distance.

\begin{figure}[H]
    \centering
    \includegraphics[width=1.0\textwidth]{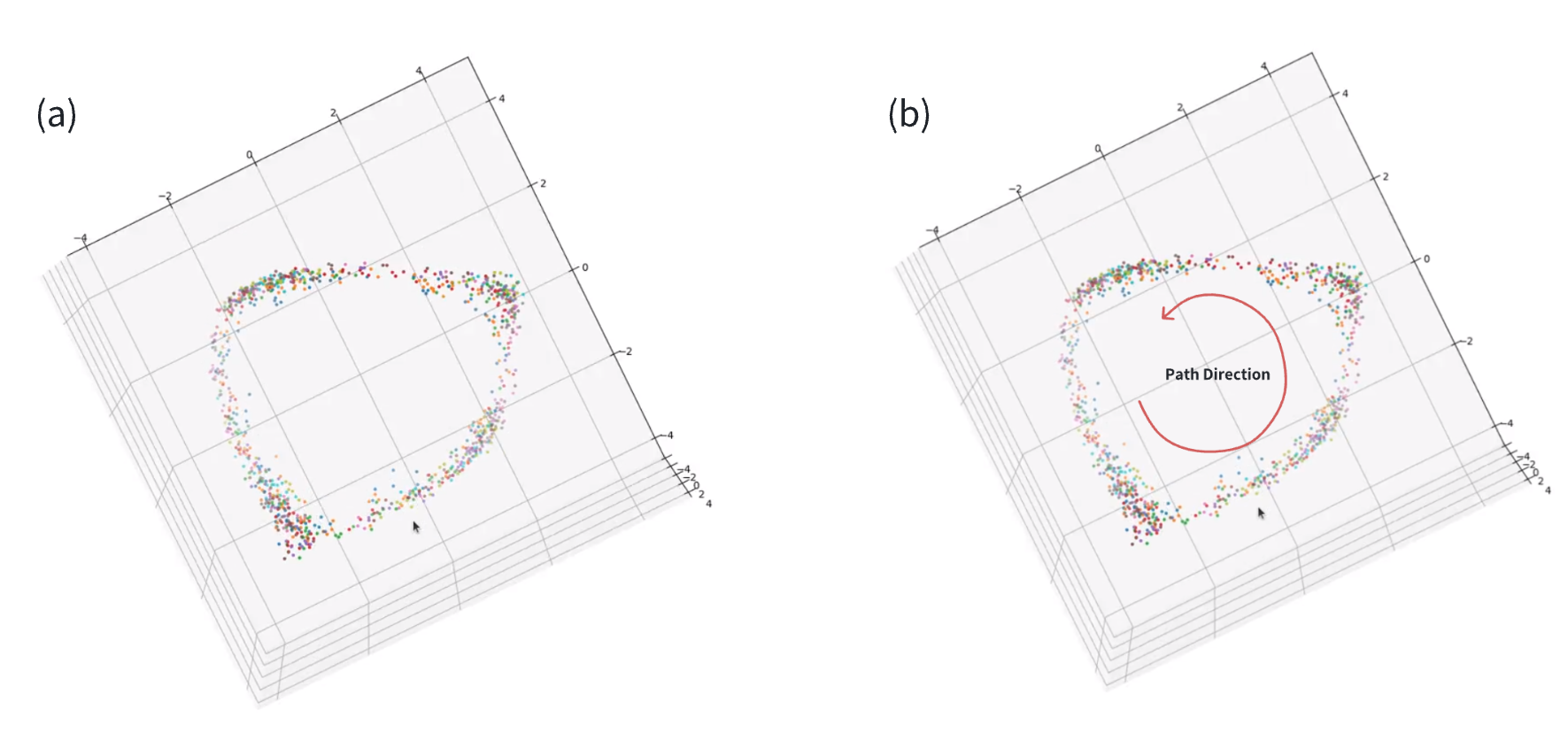}
    \caption{PCA projections revealing ring-like structures in the agent’s hidden states.
(a) A 3D scatter plot of HDS-sampled neural activations, showing a pronounced circular/elliptical distribution. Each point corresponds to a single time step in the agent’s navigation, with colors denoting different segments or trials. This “ring” indicates that the network’s internal representation systematically organizes key variables (e.g., direction or distance) around a low-dimensional cycle.
(b) The same plot with a directional cue (red arrow), underscoring how the agent’s hidden states transition smoothly along the ring as it moves through the environment. The circular shape suggests robust encoding of continuous spatial factors, reflecting an internally consistent strategy rather than rote stimulus-response mappings.}
    \label{fig:my_figure}
\end{figure}

\paragraph{Preliminary Indication}

These visual findings hint that \textbf{the network’s hidden states selectively encode direction and distance}, producing clear, separable patterns that align with navigational behavior. However, rings observed via PCA are two-dimensional abstractions and thus do not confirm whether the entire high-dimensional space aligns with spatial features or whether other geometric elements remain hidden. Similarly, we cannot yet tell how completely the network “replicates” behavioral structures based solely on the neural states.

In the next section (\textbf{2.2}), we introduce \textbf{Ridge Representation} to quantify behavior itself in a comparable manner. Ridge Representation can produce ring-shaped or polar layouts when paths differ in direction, length, and shape. Determining how strongly the ring structure in neural space couples with a similar ring in the behavioral metric space sheds further light on the agent’s depth of embodied understanding---spanning not only direction but also more complex spatial variables such as path deformation.

\subsection{Ridge Representation: A Metric-Based Ground Truth}

In the previous section (Section 2.1.3), our PCA-based analysis of HDS-sampled neural states suggested that the network encodes key spatial factors like “direction.” However, to rigorously test whether the agent truly grasps these spatial elements, we need a method to \emph{uniformly quantify navigation behaviors}—even when paths vary significantly in length or shape—and produce consistent ring-like or structured patterns on the \emph{behavioral} side. Moreover, we want to enable \emph{direct} comparisons with the high-dimensional neural data. To meet these needs, we propose \textbf{Ridge Representation}, a technique that converts two-dimensional trajectories into fixed-size grayscale images and embeds them in a “behavioral metric space.” By satisfying standard metric axioms (non-negativity, symmetry, triangle inequality) in the image domain, Ridge Representation allows subsequent one-to-one comparisons between neural activations and path images, facilitating robust analyses of their coupling.

\subsubsection{Measuring Behavioral Differences}

\paragraph{Motivation}

Different trajectories in navigation tasks vary in length or shape. Simply comparing raw coordinate sequences complicates the measurement of key attributes (e.g., direction, obstacle avoidance) and makes it difficult to run PCA, clustering, or other high-dimensional analyses on a fixed-size feature set. \textbf{Ridge Representation} addresses this by applying a linearly attenuating “radiation field” at each trajectory point and merging these point-wise effects via the “maximum value principle.” This process yields a “ridge image” (e.g., 21$\times$21 pixels) that visually encodes the path’s overall geometry and direction.

\begin{figure}[H]
    \centering
    \includegraphics[width=1.0\textwidth]{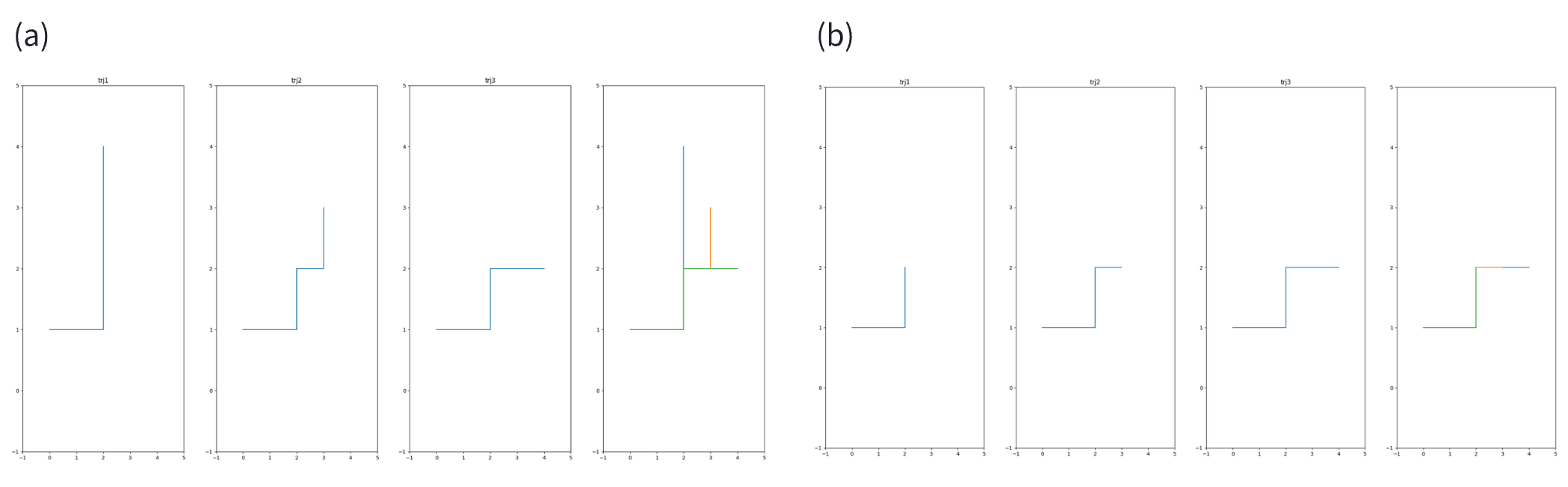}
    \caption{Motivation for a trajectory distance function that reflects path shape and overlap.
(a) Three distinct 2D trajectories (labeled trj1, trj2, trj3) that differ by both their directional patterns and endpoint positions, with a final sub-panel (far right) superimposing them for comparison. Ideally, a well-defined distance metric should yield d(trj1,trj2)<d(trj1,trj3), matching our intuitive sense that trj2 is more similar to trj1 than trj3 is.
(b) Another set of trajectories, where some segments overlap (green/orange/blue). Again, the metric should capture partial path congruence so that more overlapping routes have smaller pairwise distances. This requirement underscores the need for a robust path-distance definition that accurately compares both overall shape and local alignment.}
    \label{fig:my_figure}
\end{figure}

Because the resulting grayscale image preserves a path’s shape and orientation, we can compute Euclidean distances between images to form a behavioral feature space that fully meets metric requirements. No matter the path length or complexity, mapping all paths to identically sized images provides a uniform basis for distance calculations and low-dimensional visualizations. In other words, Ridge Representation not only standardizes the feature dimension (e.g., 441 pixels) for subsequent comparisons but also establishes a systematic metric for path differences.

\begin{figure}[H]
    \centering
    \includegraphics[width=1.0\textwidth]{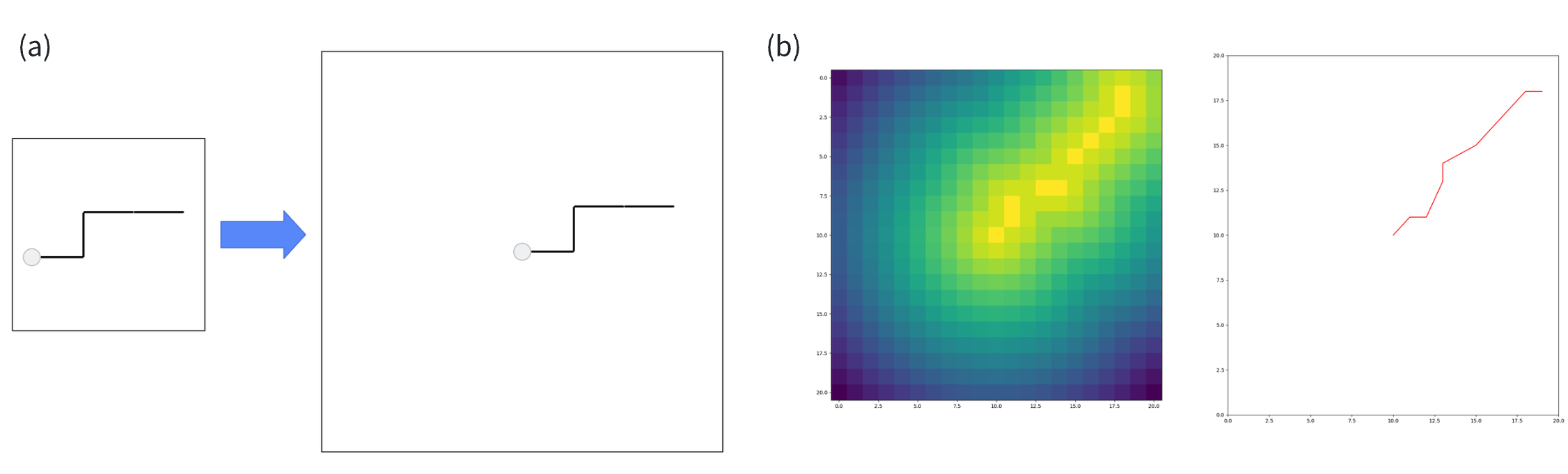}
    \caption{Illustration of the Ridge Representation generation process.
(a) A sample 2D path (left) is embedded into a larger, standardized grid (right), ensuring consistent spatial resolution regardless of original trajectory scale. The circular marker denotes the start position.
(b) The path is then converted into a grayscale “ridge image” by applying linearly decaying “radiation fields” from each point along the trajectory. The left panel in (b) shows the resulting intensity matrix (darker to brighter colors), and the right panel plots the pixel intensity along one dimension for clarity. In this representation, similar trajectories produce similar grayscale patterns, enabling a uniform and intuitive distance metric that quantifies path shape and alignment.}
    \label{fig:my_figure}
\end{figure}

\paragraph{Ring-Like PCA in Ridge Space}

After converting thousands of navigation paths into ridge images, we can perform PCA on these images, just as we did for neural states. Indeed, when the path direction shifts gradually (e.g., from “up” to “right” to “down”), the PCA projection of ridge images often forms an \emph{approximately ring-shaped} distribution:

\begin{itemize}
    \item Angles around the circle correspond to different path directions.
    \item The radius or outward gradient from the circle’s center often reflects path length or more subtle deformations.
\end{itemize}

Such ring-shaped distributions closely resemble those observed in the PCA of neural states, suggesting parallel metric structures in higher-dimensional space. \emph{If the ring in neural space aligns with the ring in Ridge space across multiple dimensions, then the network has effectively “mirrored” the geometric relationships of the behavioral space in its activations.} Conversely, if any perceived similarity appears only in a few low-dimensional projections while diverging in most dimensions, the resemblance could be superficial.

\begin{figure}[H]
    \centering
    \includegraphics[width=0.8\textwidth]{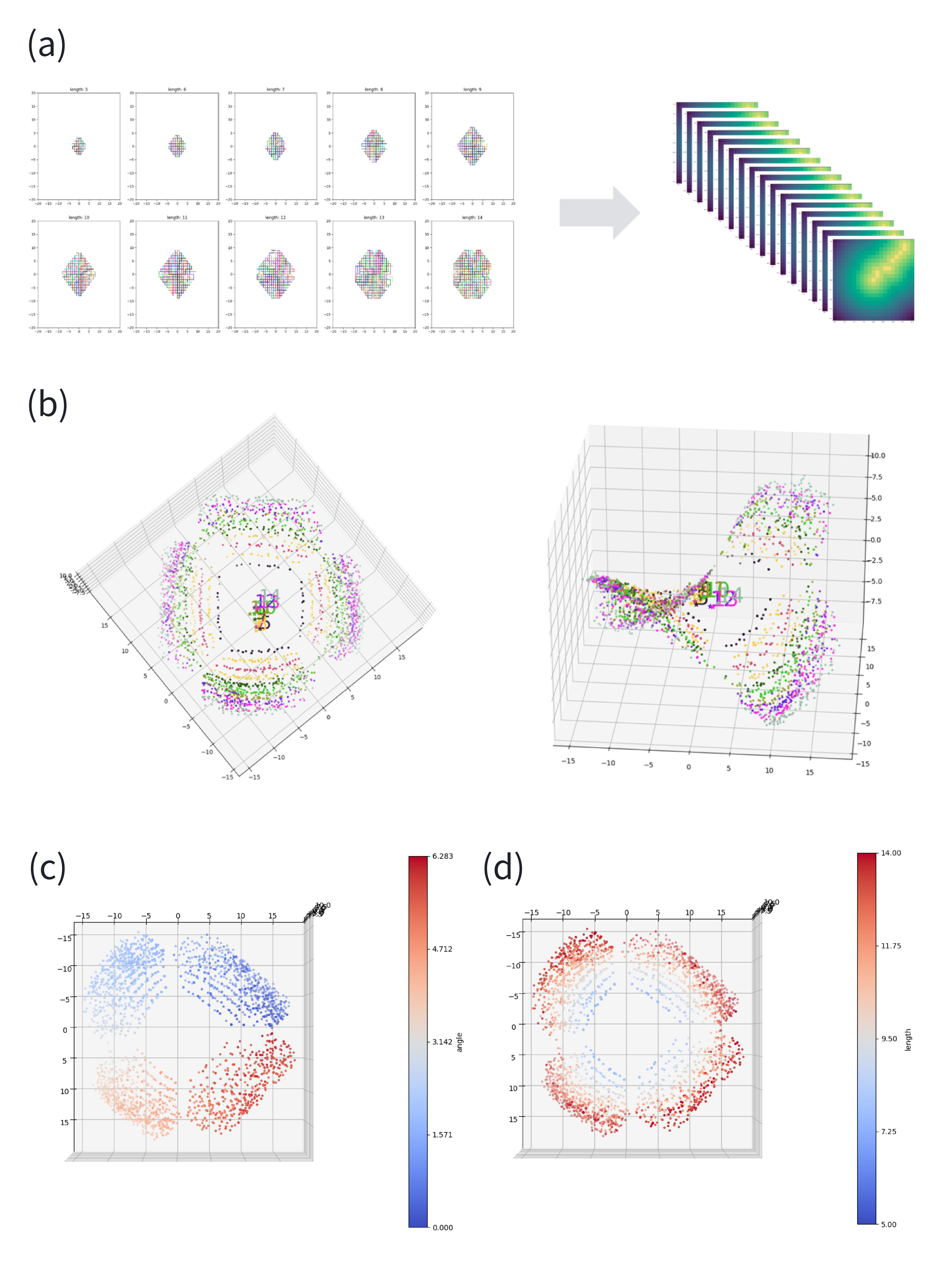}
    \caption{Illustration of the Ridge Representation generation process.
(a) A sample 2D path (left) is embedded into a larger, standardized grid (right), ensuring consistent spatial resolution regardless of original trajectory scale. The circular marker denotes the start position.
(b) The path is then converted into a grayscale “ridge image” by applying linearly decaying “radiation fields” from each point along the trajectory. The left panel in (b) shows the resulting intensity matrix (darker to brighter colors), and the right panel plots the pixel intensity along one dimension for clarity. In this representation, similar trajectories produce similar grayscale patterns, enabling a uniform and intuitive distance metric that quantifies path shape and alignment.}
    \label{fig:my_figure}
\end{figure}

\subsubsection{Structural Similarity \& Open Questions}

\paragraph{Direction vs. Other Features}

In relatively simple path scenarios, direction is the most prominent geometric factor and is easily visible in PCA projections. Ridge-based PCA clearly distinguishes directions like “up/right/down/left,” while also reflecting path length or turning points. However, \emph{length and finer spatial variations} may not stand out in principal components, as they can be overshadowed by major directional differences. To confirm whether the network truly encodes these subtle features—like obstacle avoidance or detailed path deformations—\textbf{high-dimensional} neural-behavioral comparisons are needed.

\begin{figure}[H]
    \centering
    \includegraphics[width=1.0\textwidth]{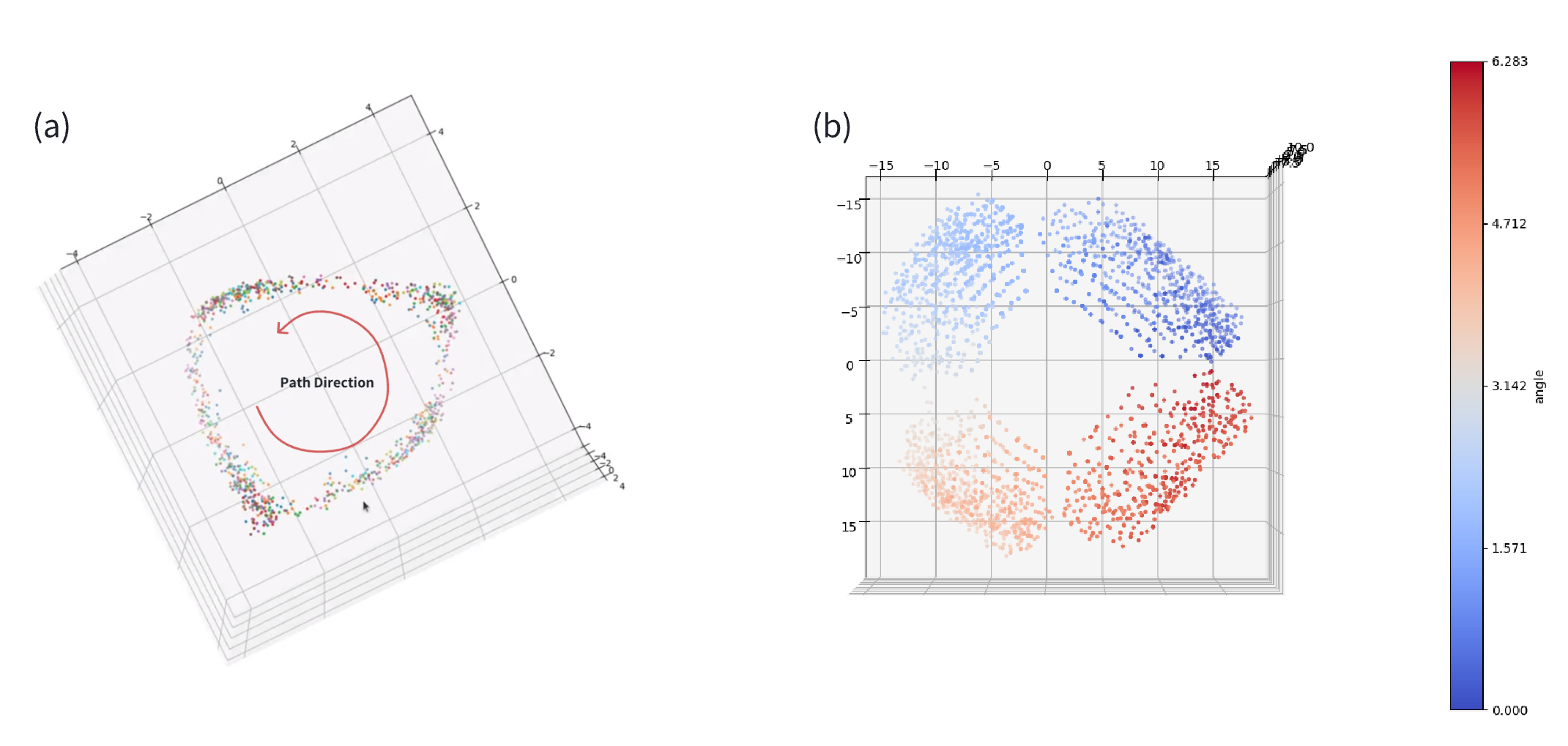}
    \caption{Comparing directional versus other feature encodings in Ridge-space PCA.
(a) A 3D PCA projection of Ridge-embedded trajectories, forming a pronounced ring structure. The red arrow indicates how path direction traverses the ring smoothly, highlighting direction as a dominant spatial factor in the embedding.
(b) Points color-coded by trajectory angle (from blue to red) further reveal how continuous changes in direction correspond to distinct arcs around the ring. Although direction is the most salient variable, additional geometric features (e.g., path shape, distance) may also appear as subtle variations within this low-dimensional manifold.}
    \label{fig:my_figure}
\end{figure}

\paragraph{Need for High-Dimensional Analysis}

Although PCA suggests both neural states and Ridge images may form rings related to path direction, this remains a \emph{qualitative or low-dimensional} observation. We still lack a quantitative assessment of how strongly they couple in the full high-dimensional space. To determine whether neural activations encode more nuanced aspects (e.g., path deformation, small-scale turns), we need an approach that directly compares 128-dimensional neural vectors to 441-dimensional Ridge images. Enter \textbf{Canonical Correlation Analysis (CCA)}, which uncovers multiple pairs of projection directions maximizing correlation across the two spaces. \emph{If CCA yields high correlation coefficients for the first 5–10 projection pairs}, we can infer that the agent’s internal replication of the behavioral space is both broad and robust.

Still, correlation alone does not imply causation for decision-making. Hence, in Section~2.3.3, we describe “key-dimension intervention” experiments that examine the causal impact of highly correlated dimensions on navigation. This progression—from demonstrating statistical coupling to confirming actual influence on agent behavior—lets us move beyond \emph{mere correlation} toward causal evidence of “what truly drives navigation decisions.”

\subsection{Quantifying Neural-Behavior Correlation via CCA}

In Section~2.2, we showed that \textbf{Ridge Representation} produces ring structures similar to those observed in neural states (via PCA), suggesting an underlying alignment between the agent’s high-dimensional neural activations and its behavioral metrics. However, low-dimensional visualizations alone cannot prove that this alignment extends across \emph{all} dimensions or confirm that it reflects a systematic correspondence rather than coincidence. To address this, we introduce \textbf{Canonical Correlation Analysis (CCA)}, a method that identifies multiple pairs of optimal projection directions between two high-dimensional datasets—in this case, the neural states and Ridge-based behavioral features. By examining whether these projections show significant correlations, we can better determine whether the agent’s neural representation replicates the geometry of the behavioral metric space. We then verify through intervention experiments whether these dimensions directly influence decision-making rather than merely correlating with behavior.

\subsubsection{CCA Setup \& Rationale}

\paragraph{Direct High-Dimensional Mapping}

Before applying CCA, we did not reduce the dimensionality of either dataset: the neural states (generally 128 dimensions) or the Ridge images (often 441 dimensions). Instead, we simply performed centering and standardization, then supplied them directly to the CCA algorithm. This preserves as much underlying geometric and directional information as possible and avoids discarding potentially important features.

Intuitively, if the agent has truly internalized multiple spatial factors—such as direction, obstacle avoidance, and small-scale path deformation—there should be several projection axes in the high-dimensional space along which neural states and Ridge descriptors exhibit strong linear coupling. In contrast, if only one or two such projections show modest correlation, it would suggest that the agent has only captured partial elements (e.g., rough direction) without forming a deeper, more systematic spatial representation.

\paragraph{Core Hypothesis}

Mathematically, CCA finds sets of linear projection vectors $\mathbf{a}_i$ and $\mathbf{b}_i$ that map $\mathbb{R}^{N \times 128}$ (neural data) and $\mathbb{R}^{N \times 441}$ (Ridge data) into low-dimensional subspaces where correlation is maximized. Our hypothesis is that if the agent robustly encodes spatial attributes crucial for navigation, significant correlations $(\rho_i \approx 0.8 \text{ to } 0.9)$ should emerge in the first several (e.g., 5–10) projection pairs. Such multi-dimensional correlations would indicate a broad, consistent alignment, implying strong internalization of spatial structure. Conversely, if only one or two projection pairs display small correlations, it suggests the network’s encoding is shallow.

\subsubsection{CCA Results \& Baseline Comparison}

\paragraph{High Correlation Across Multiple Directions}

After consolidating the neural and Ridge feature matrices, we computed canonical correlation vectors using standard techniques (e.g., singular value decomposition). In the first 5–10 projection pairs, the correlation coefficients $\rho_i$ consistently reached 0.8–0.9 or above, well beyond random noise levels. This indicates a broad and stable alignment between neural activations and behavioral metrics across multiple directional subspaces.

Further inspection revealed that pixel regions in Ridge images most correlated with “direction” or “obstacle location” matched noticeable gradients or blocks in the corresponding neural projections. This suggests that the network encodes not just coarse directional cues but also finer spatial details, reinforcing the claim that neural representations extend beyond simple memorization or macro-level patterns.

\begin{figure}[H]
    \centering
    \includegraphics[width=0.8\textwidth]{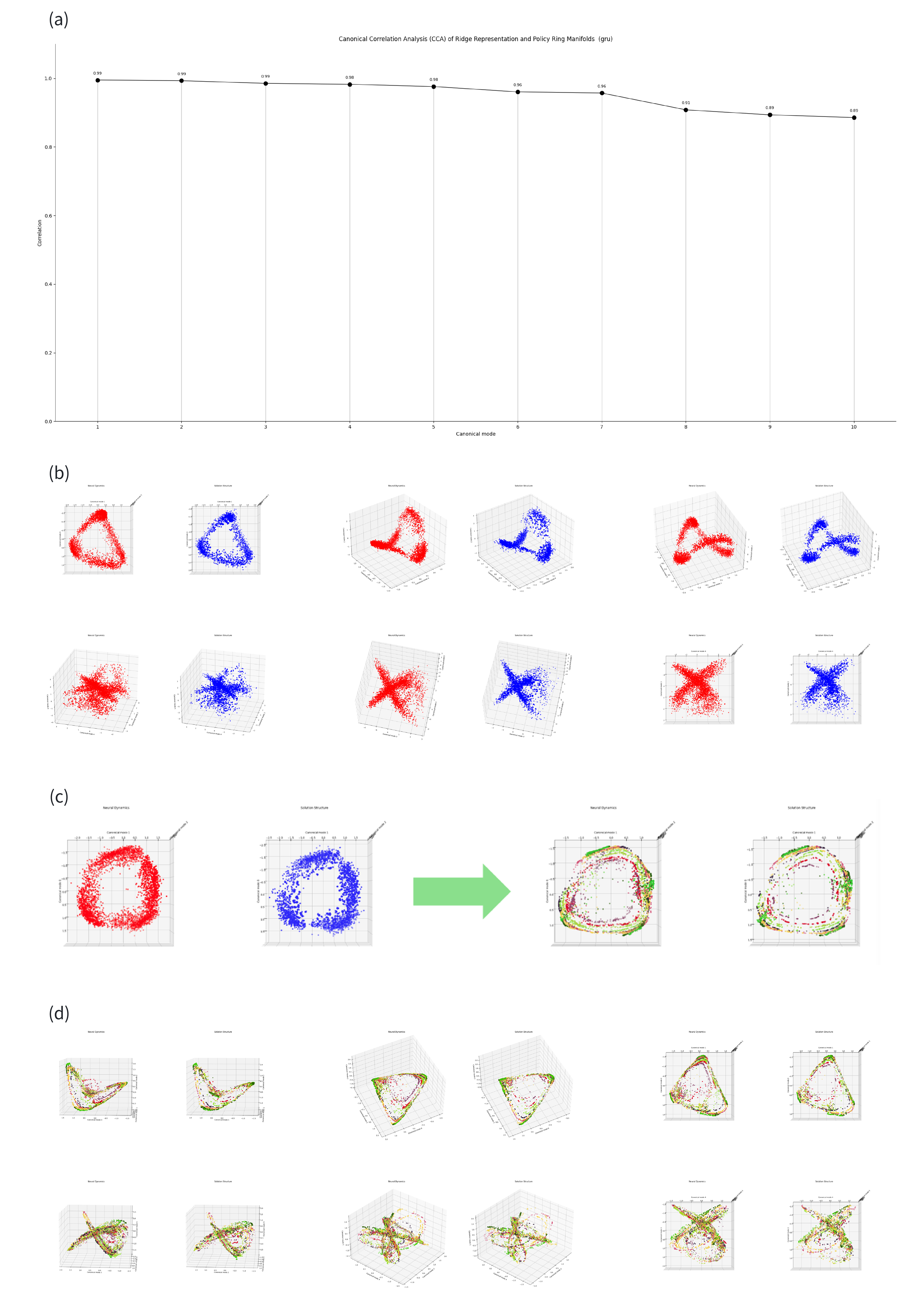}
    \caption{Canonical Correlation Analysis (CCA) revealing high-dimensional alignment between neural and behavioral data.
(a) Correlation coefficients for the first 10 canonical modes, showing consistently high values, indicating a strong linear relationship between neural embeddings and the Ridge-based behavioral features.
(b) After projecting both datasets into their shared CCA space, the neural data (red) and behavioral data (blue) closely overlap under multiple viewing angles. The top row visualizes modes CM0–2, while the bottom row shows CM3–5. Each sub-figure confirms that points from the two modalities form nearly identical structures in the canonical subspace, illustrating robust alignment beyond mere low-dimensional coincidence.
(c) To examine finer-grained structure, data are grouped by path length and averaged within local neighborhoods; the color gradient reflects increasing path length. Superimposing these clusters onto the canonical modes reveals detailed correspondences between neural states and behavioral paths, preserving both global and local features.
(d) Multiple viewpoint snapshots of the refined clusters in CM0–2 (top row) and CM3–5 (bottom row). Even at different orientations, the red (neural) and green (behavioral) point clouds demonstrate near-exact matching, underscoring the network’s capacity to encode nuanced path geometry in a high-dimensional yet systematically shared latent space.}
    \label{fig:my_figure}
\end{figure}

\paragraph{Cross-Validation \& Random Baseline}

To rule out spurious correlations or artifacts tied to particular maze setups, we repeated the same procedure across various maze configurations, training phases, and cyclic stimulation conditions. The strong multi-dimensional correlation consistently appeared, indicating that this alignment was neither scenario-specific nor incidental. Additionally, when we applied the same procedure to random networks or models retaining only the output layer, correlation coefficients stayed near 0.1–0.2. This stark contrast demonstrates that high correlation arises specifically in agents that genuinely learned spatial features.

\begin{figure}[H]
    \centering
    \includegraphics[width=1.0\textwidth]{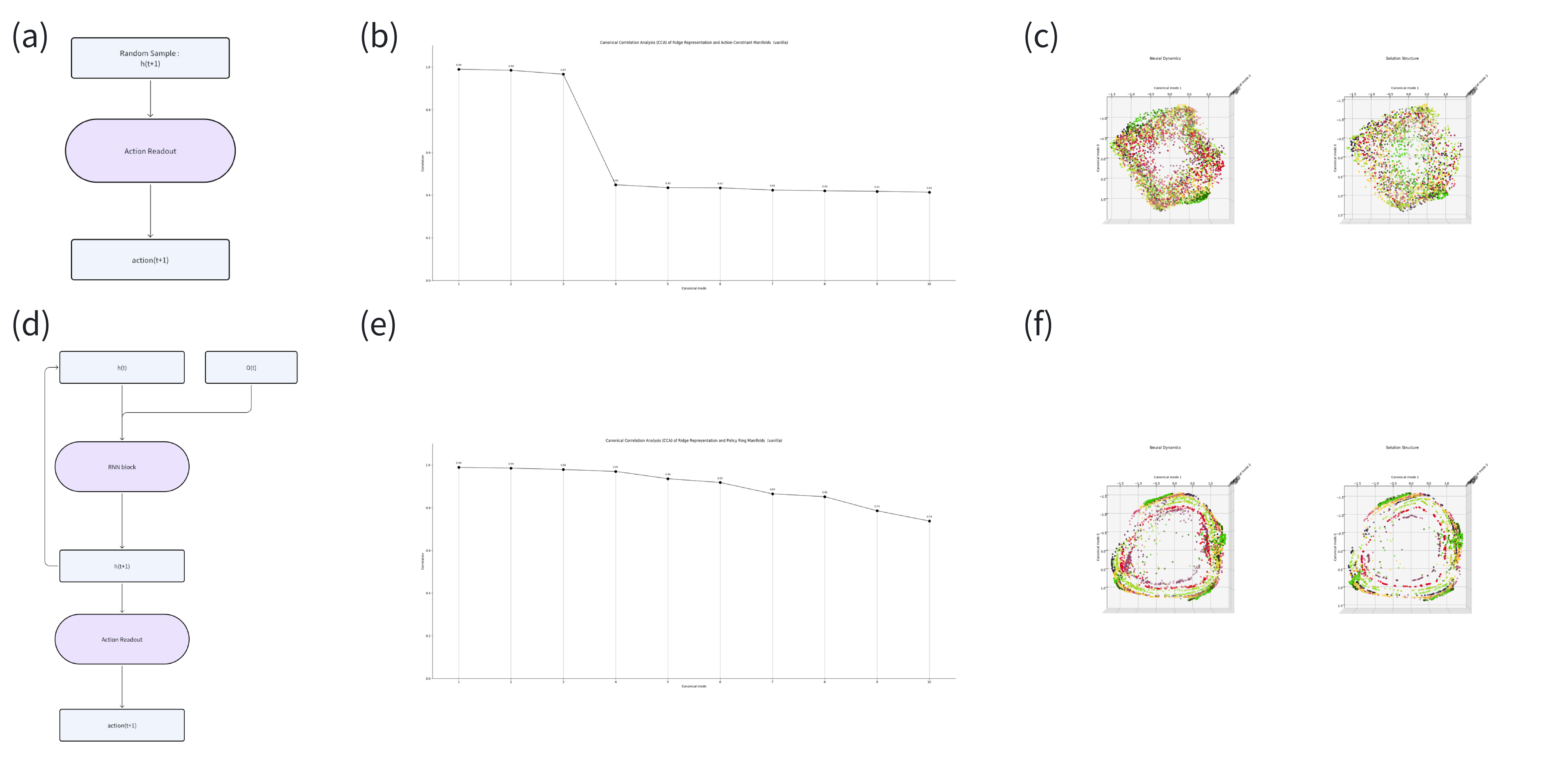}
    \caption{Cross-validation and random baseline experiments underscoring the significance of learned representations.
(a) Schematic of the random baseline network, where hidden states are partly constrained to produce specific actions while other dimensions are randomized. This setup artificially links some action dimensions to the environment but leaves most neural coordinates unstructured.
(b) CCA results for the random network vs. Ridge-based paths, showing strong correlation in only the first few modes and a rapid drop from CM3 onward, characteristic of noise-like latent variables that do not systematically align with behavior.
(c) 3D CCA projections for the random network’s neural (red) and behavioral (green) data. While coarse global structures might overlap, finer local patterns are absent, indicating the lack of true feature correspondence.
(d) Flow diagram of the standard (trained) RNN-based agent, integrating environment observations (obs(t)) and hidden states (h(t)h(t)) to produce actions.
(e) In contrast to the random baseline, the trained agent’s CCA correlations remain high across multiple modes, reflecting robust alignment between neural and behavioral representations.
(f) 3D visualizations in canonical space, showing near-perfect overlap of neural (red) and Ridge-based (green) point clouds with detailed local matching, demonstrating a deeply learned, high-dimensional correspondence that goes far beyond the macro scale.}
    \label{fig:my_figure}
\end{figure}

\paragraph{Implications}

By this stage, we can reasonably conclude that \emph{the agent’s network exhibits robust linear correspondences with the geometric properties of navigation paths}, corroborating the hints from PCA-based analyses. Nonetheless, statistical correlation alone does not prove that these neural dimensions drive decision-making. Potentially, they could merely track behavior without directing it. To address this causal question, we designed intervention experiments in Section~2.3.3, focusing on \textbf{the most highly correlated dimensions}. If disrupting these dimensions immediately impairs navigation, it strongly indicates they are integral to the final strategy, rather than coincidental byproducts.

\subsubsection{Control \& Counterfactual Intervention}

Here, we evaluate whether deliberately disrupting the neural dimensions identified by CCA has a measurable impact on navigation. If performance deteriorates under such interventions, it confirms that these dimensions hold a causal role in encoding spatial information for direction and obstacle avoidance.

\paragraph{Why Causal Testing}

Even when correlations reach $(\rho_i \approx 0.9)$, it is still possible that neural activations passively fluctuate with behavior. Only by imposing targeted interventions—such as zeroing out or randomizing specific dimensions—and observing significant performance decline can we confirm these dimensions actively guide navigation, refuting purely “post-hoc” or “pseudo-correlation” explanations.

\paragraph{Dimension Selection \& Manipulation}

We typically select the top 5–10 projection dimensions from CCA (or those with the largest variances or gradients within the neural activation space) and intervene by zeroing their values or overwriting them with random noise. The rest of the neural dimensions remain unchanged. We then measure path length, success rate, or other metrics under the same maze settings (and cyclic stimulation sequences, if used) to see if performance degrades.

\paragraph{Observed Performance Improvement}

Empirical results show that injecting the high-correlation dimensions identified by CCA—namely, substituting the agent’s current hidden state with the neural signals tied to the corresponding physical trajectory—significantly boosts navigation efficiency. Path lengths are reduced, success rates increase, and the agent converges more rapidly on optimal strategies, reflecting the strong causal contribution of these dimensions. Conversely, providing random or low-correlation dimension signals yields negligible performance gains, confirming that the CCA-aligned dimensions indeed capture critical navigational features.

\paragraph{Counterfactual Reasoning}

To exclude alternative interpretations, we examined multiple scenarios:

\begin{enumerate}
    \item \textbf{Preserving only the high-correlation dimensions} and randomizing the others. Agents still navigate relatively well.
    \item \textbf{Randomizing high-correlation dimensions} while preserving the others. Agent performance collapses.
\end{enumerate}

Such dramatic contrasts show that only those dimensions strongly coupled with the Ridge representations (via CCA) encode direction/obstacle data essential for successful navigation.

\begin{figure}[H]
    \centering
    \includegraphics[width=1.0\textwidth]{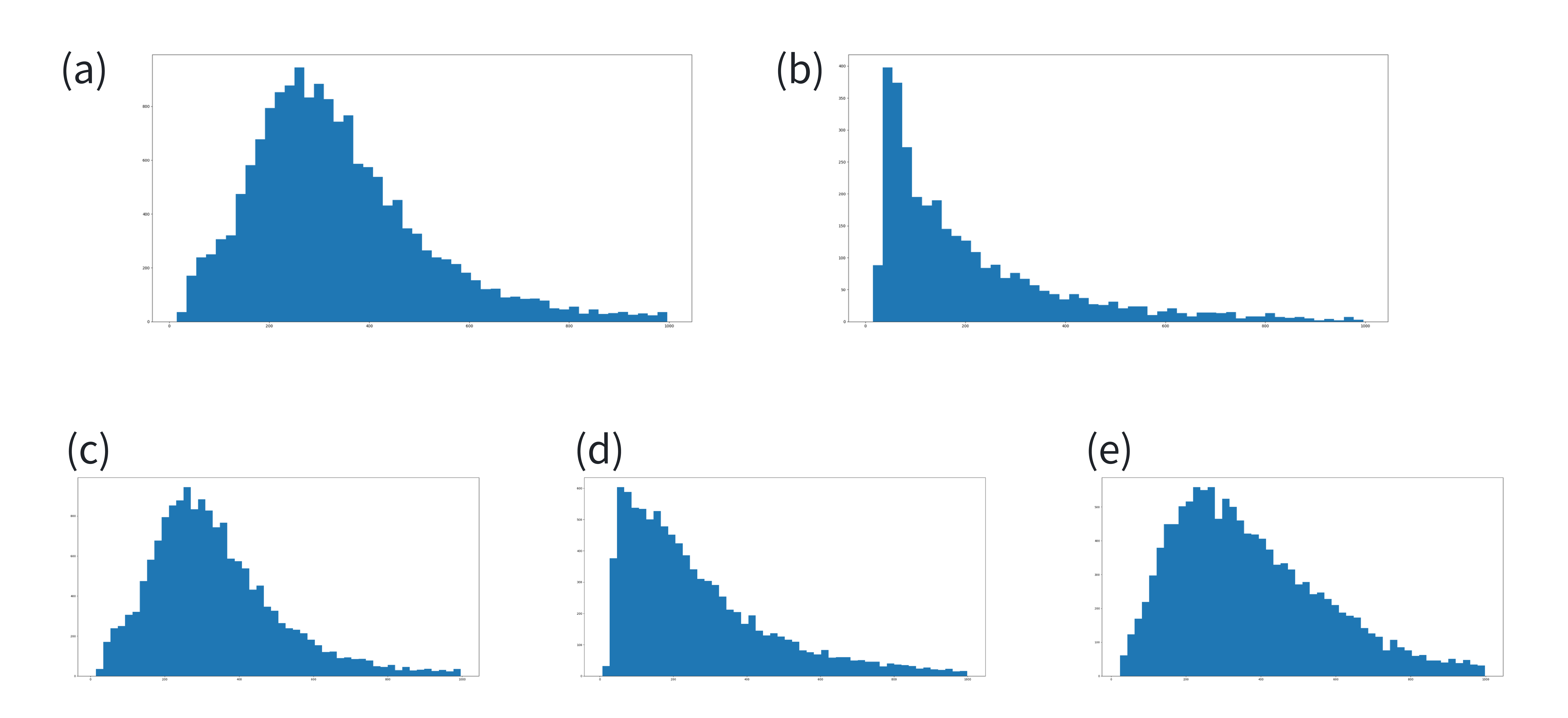}
    \caption{Histograms of convergence times under different neural-injection and dimension-randomization conditions.
(a, c) In the baseline scenario (normal run), the distribution of path convergence times peaks around 250 steps, indicating the typical speed at which agents stabilize on a successful route.
(b) When the agent is provided with “optimal strategy” neural representations—directly injecting the hidden states that correspond to the best-performing physical trajectory—the convergence histogram shifts dramatically to a peak near 50 steps, showing faster stabilization and higher efficiency.
(d) Even if all remaining dimensions are randomized, preserving only the optimal-strategy dimensions retains a similar accelerated convergence peak (around 50 steps), underscoring the causal importance of these high-correlation dimensions.
(e) By contrast, randomizing the optimal-strategy dimensions while keeping all other neural components intact reverts the convergence peak to around 250 steps, further confirming that the injected dimensions are key drivers of navigation success.}
    \label{fig:my_figure}
\end{figure}

\paragraph{Implications}

These intervention studies confirm the causal importance of the highly correlated dimensions: they do not merely co-occur with certain behaviors but \emph{drive} the agent’s navigation decisions. Together, \textbf{statistical correlation plus intervention-driven performance decline} establishes convincing evidence that the network internalizes spatial concepts (direction, distance) in a deeply integrated manner. Within the HDS framework and using Ridge Representation as introduced here, the agent’s hidden states not only reflect essential spatial factors (direction, path deformation) but also determine the agent’s near-optimal decision strategy. This finding provides further support for the feasibility of \textbf{embodied cognition} in artificial neural networks.

\section{Discussion}

This study explores whether neural networks can spontaneously acquire deep spatial understanding—encompassing direction, distance, and obstacle-avoidance strategies—through continuous interaction with the environment, focusing on \textbf{planar navigation tasks} within an \textbf{embodied cognition} framework. By combining \textbf{Hybrid Dynamical Systems (HDS)} modeling with \textbf{Ridge Representation} metrics for behavior, we have empirically demonstrated that agents not only achieve efficient navigation but also exhibit neural activations closely aligned with key features of the behavioral path space. Below, we discuss the theoretical and practical significance of these findings in relation to embodied cognition and explainable AI, as well as current limitations and potential directions for future research.

\subsection{Implications for Embodied Cognition}

\emph{This work supports a long-standing hypothesis in embodied cognition}: when agents learn through repeated physical interaction with their surroundings, their internal representations extend beyond the superficial “see target $\to$ move toward target” paradigm to encode higher-order geometric concepts such as direction and distance. Our findings show that not only do agents plan near-optimal paths, but their neural states also \textbf{explicitly} capture these spatial elements—manifested as ring-like or gradient patterns indicating directional segmentation and path deformation. Coupled with intervention experiments, we confirm that these representations are not merely byproducts but actively influence the agent’s choices.

From the lens of embodied cognition, this underscores that \textbf{repeated perception–action coupling} is crucial for “understanding.” Agents must continually act upon their environment and update internal models based on new states to establish deep spatial comprehension. Passive observation alone typically fails to produce robust attractor-like structures (e.g., limit cycles). Mastering concepts such as “direction/distance” thus goes beyond simple stimulus-response mappings, emerging as a \emph{systematic encoding of concept relationships} (including obstacle avoidance and local detours). This echoes the view that “comprehension arises from action–perception loops”: as agents “walk,” they shape their internal dynamical systems to handle goal reaching and obstacle avoidance in diverse settings.

\subsection{Hybrid Dynamical Systems Analysis and Resolving the “Causal Paradox”}

A major innovation of this study is our use of \textbf{HDS (Hybrid Dynamical Systems)}, which integrates agent–environment coupling into a fully closed dynamical model. We further introduce \textbf{cyclic stimulation} and \textbf{Lyapunov exponents} to identify stable attractors representing optimal strategies in the hybrid space. These techniques address the common “causal paradox” in traditional explainable AI for embodied scenarios: when network activations both respond to and shape the environment, discerning cause from effect can be challenging. By modeling the combined system $(Q, X)$ and checking for \textbf{stable limit cycles} or \textbf{attraction under perturbations}, we find that agents embed optimal action loops deeply within a unified, closed-loop dynamics—rather than simply reacting to external inputs or enforcing one-sided control.

The \textbf{cyclic stimulation} technique functions as a “minimal external intervention” that momentarily blocks real environmental feedback. If the network then re-enters and remains in the same periodic attractor identified earlier (using only a fixed observation sequence), it demonstrates that this loop is not only observable in standard interactions but also robust to noise or initial-state variations. This finding holds broader significance. In many robotics and continuous-control tasks, where the environment and agent can be jointly expressed as a dynamical system, one can apply cyclic stimulation to detect whether agents have internalized self-consistent attractors—overcoming the one-directional notion of “network $\to$ action” and clarifying causal loops like “action $\to$ environment $\to$ network.”

\subsection{Limitations and Future Outlook}

Despite meaningful progress in both empirical and methodological domains, several constraints remain to be addressed in future work:

\begin{enumerate}
    \item \textbf{2D Discrete Maze Environment}

    Our setting features relatively small-scale grids with simplified perception and actions. In higher-dimensional or continuous navigation scenarios (e.g., real-world robotics, three-dimensional path planning), agents may develop more varied limit cycles or multi-stable attractors. Whether HDS + Ridge + intervention methods apply equally well in such complex domains merits further exploration.

    \item \textbf{Network Architecture Diversity}

    This study focuses on recurrent networks (GRUs) and does not systematically compare other architectures (e.g., LSTM, Transformer encoders). If more sophisticated models display similar spatial encoding, it may suggest that embodied representations are a broadly shared property; if different designs produce distinctive patterns of encoding, they may offer unique advantages for specific embodied cognition tasks.

    \item \textbf{Experiment Scale}

    While we conducted thousands of maze trials, the total data volume and range of scenarios could still grow. More dynamic tasks—such as those with moving obstacles or time-varying goals—might yield additional attractors or more intricate hybrid state structures. Extending training runs might also further stabilize or enhance the encoding of secondary spatial features (e.g., subtle path variations).
\end{enumerate}

Nevertheless, this study provides a solid foundation for understanding how neural networks acquire embodied world models through physical interaction. Future investigations might expand these techniques to complex continuous-control domains, implement them in physical robot platforms, or explore how they relate to broader agent capabilities (e.g., semantic reasoning, social interaction). By integrating \textbf{HDS} modeling, \textbf{Ridge Representation}, and \textbf{CCA-based intervention}, we open new avenues for analyzing embodied intelligence across a wide range of applications.

\medskip
\hrule
\medskip

\section{Methods}

This section describes our experimental design, network architecture, and analytical techniques. In \textbf{Section~4.1}, we outline the planar navigation environment and task setup. \textbf{Section~4.2} covers the agent’s neural architecture and meta-reinforcement learning framework. \textbf{Section~4.3} introduces the Hybrid Dynamical Systems (HDS) method for modeling agent–environment interactions, and \textbf{Section~4.4} details the Ridge Representation for encoding behavioral features. Finally, \textbf{Sections~4.5} and \textbf{4.6} explain our high-dimensional correlation analyses using Canonical Correlation Analysis (CCA) and intervention experiments to test causal relationships.

\subsection{Task \& Dataset}

Our central navigation task is an \textbf{Open-Ended random planar maze}, implemented as a 10$\times$10 discrete grid. During maze generation, each cell—except for the start (green) and goal (red)—is assigned an obstacle with about 30\% probability via uniform random distribution. Afterward, we check connectivity (e.g., via OpenCV’s connected component analysis) to discard any unsolvable layouts. Although nominally a 10$\times$10 grid, there are $2^{100}$ possible obstacle configurations if connectivity is not considered. Even at a fixed 30\% obstacle density, this massive search space prevents the agent from relying on memorized scenarios or exhaustively enumerating all configurations. Instead, it must learn more generalizable internal models to cope with such diversity in both training and testing.

\begin{figure}[H]
    \centering
    \includegraphics[width=1.0\textwidth]{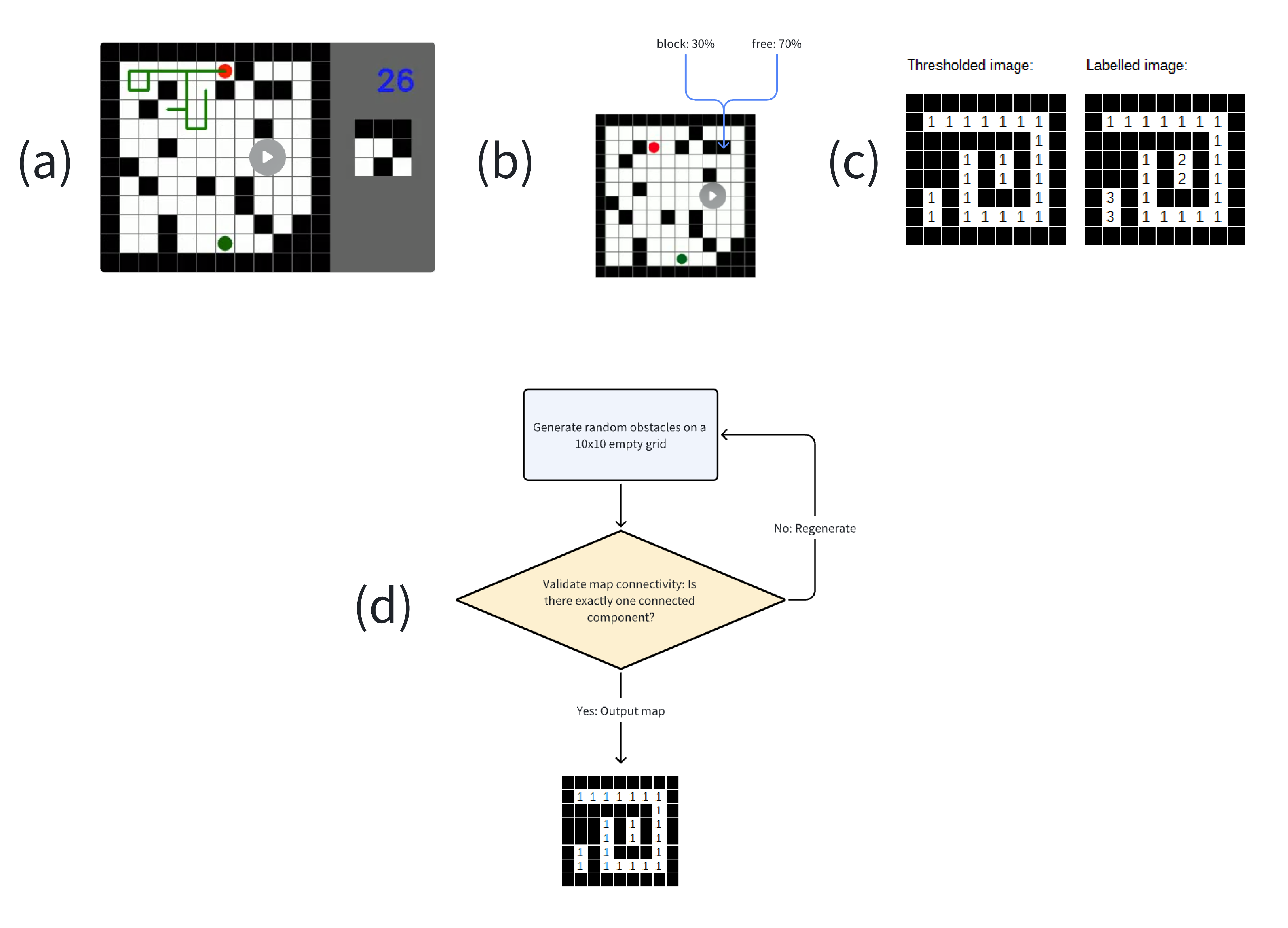}
    \caption{Randomized terrain generation and partial-observation setup for the navigation task.
(a) The agent’s 3×3 local perception window (green squares), illustrating how it only sees tiles immediately surrounding its current grid cell. This mimics real-world sensory constraints by limiting the agent’s global awareness.
(b) Obstacles (black squares) are randomly placed across the 10×10 grid at a density of roughly 30
(c) To ensure valid navigation, each generated grid is passed to an OpenCV-based connected component algorithm. Grids containing more than one connected region are discarded and regenerated.
(d) Flowchart of the terrain generation pipeline: the system repeatedly samples random obstacle configurations and checks connectivity. Only maps with exactly one connected component are retained, guaranteeing that the agent can traverse from start to goal.}
\end{figure}

Within this grid, the agent has \textbf{local perception} restricted to a 3$\times$3 neighborhood centered on its current location. If adjacent cells lie outside the grid or contain obstacles, those corresponding entries in the perception array become null or obstacle labels. This limited “field of view” represents a pronounced \emph{information bottleneck}: the agent cannot see the entire maze at once, so it must continuously explore and build a mental model of the maze’s layout over time. Each new observation depends on the previous action, forming a tight “perception–action” loop that requires a network with internal memory (e.g., GRU) to store and combine past observations. Only after accumulating enough temporal information can the agent devise near-optimal or fully optimal paths. This necessity of integrating perception and action over time aligns with the embodied cognition theory, suggesting that such training environments foster spontaneous formation of \textbf{deep spatial–behavioral representations}.

The agent chooses among four discrete actions—\textbf{up, down, left, right}—each time step. Attempting to move to an adjacent cell succeeds if that cell is empty; if it is blocked or out of bounds, the move is invalid (the agent either stays put or the action is effectively ignored). This discrete action space balances simplicity and interpretability. When the agent reaches the goal, navigation is deemed successful, and only its physical position resets to the start; its internal neural state remains \emph{intact}. As a result, the agent “remembers” the maze from earlier explorations and can optimize subsequent paths or attempt new strategies. Without this continuity of internal memory, the agent would need to relearn the layout from scratch after every successful attempt, significantly hampering improvement in the current maze.

\begin{figure}[H]
    \centering
    \includegraphics[width=1.0\textwidth]{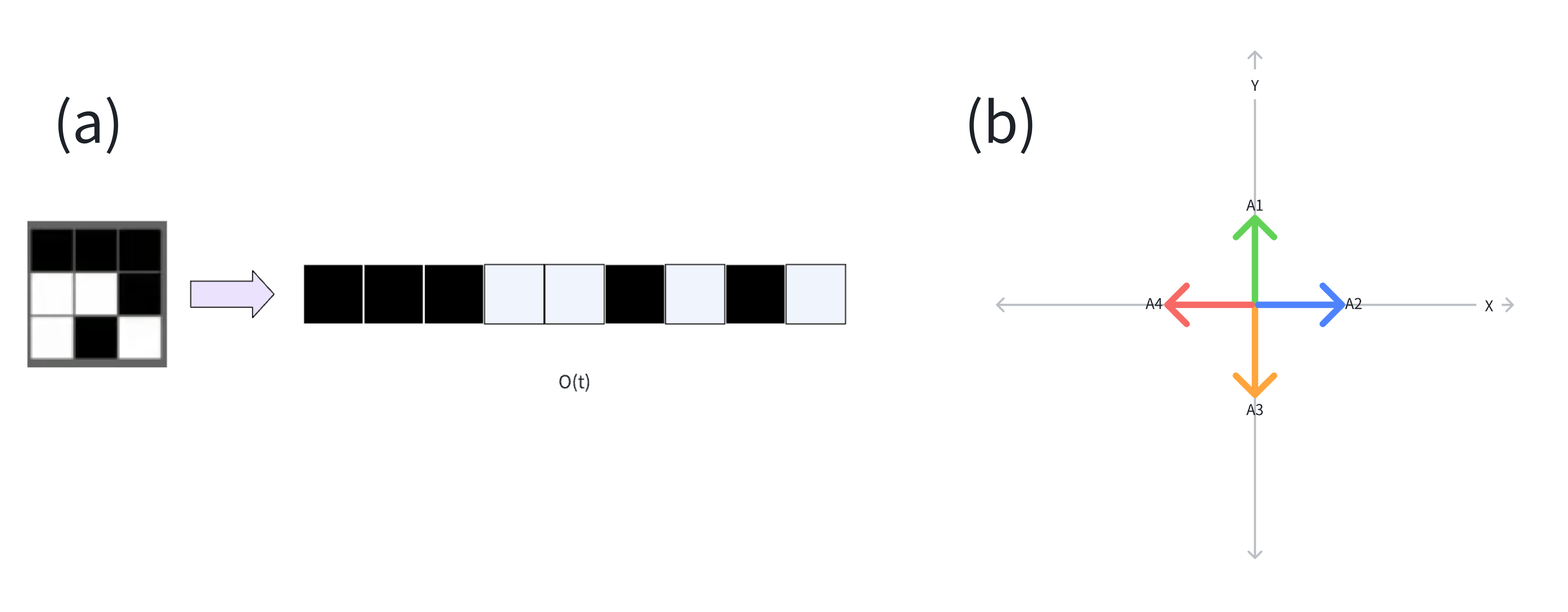}
    \caption{Observations and action space for the planar navigation agent.
(a) Each 3×3 local patch around the agent’s current location is flattened into a one-dimensional vector O(t)  for network input. Black squares indicate obstacles, while white or gray squares are traversable cells. This design imposes partial observability, compelling the agent to rely on internal memory.
(b) The agent’s action space comprises four discrete moves—up (A1), right (A2), down (A3), and left (A4)—representing single-step movements in the corresponding cardinal directions.}
\end{figure}

In this \textbf{meta-reinforcement learning (meta-RL)} framework, the agent typically makes multiple attempts in the same maze. We quantify performance by the \textbf{path length on the final reset}—if the agent effectively internalizes start and goal positions and identifies useful routes, the last path should approach optimality; if not, it remains unnecessarily long. During training, we apply a center-rank transformation to these path lengths for fitness assignment: agents that constantly take detours or hit obstacles accumulate lower fitness scores and thus less influence on subsequent evolution.

Overall, our use of open-ended random mazes, combined with local perception and a discrete action space, creates a challenging scenario demanding multi-step exploration and persistent memory. The reliance on repeated trial-and-error to develop robust spatial models underscores the benefits of embodied cognition principles. Subsequent sections detail how we design and train the agent’s network and how we analyze the resulting behavioral and neural dynamics.

\subsection{Agent Architecture \& Training}

To enable effective learning in locally observable maze environments with repeated exploratory episodes, we made careful design choices at both the network and training-strategy levels. Below, we describe the neural network (including its input and gating mechanisms), details of the perception–action cycle, the fitness evaluation and evolutionary scheme, and finally the meta-reinforcement learning (meta-RL) setup that allows rapid adaptation to new mazes.

\subsubsection*{Neural Network Implementation}

At each time step, the agent concatenates the current 3$\times$3 local observation with its previous hidden state $h_{t-1}$ before feeding this combined vector into the network. This approach lets the network incorporate the latest observations while retaining a memory of local maze structure. The concatenated input first enters a \textbf{Gated Recurrent Unit (GRU)}, which regulates memory retention through reset and update gates. Specifically, the GRU calculates reset gate $r_t$ and update gate $z_t$ from the current input $x_t$ and the previous hidden state $h_{t-1}$, then combines them with a candidate state $\hat{h}_t$ to yield the new hidden state $h_t$. In a setting with limited local perception, the GRU’s gating mechanism helps the network retain critical information—such as crucial routes or environmental cues—while discarding noise, thus facilitating more efficient navigation.

Once the GRU outputs the new hidden state $h_t$, a linear (fully connected) layer maps it to logits for the four possible actions (up, down, left, right), which are transformed via softmax into a probability distribution $\pi(a \mid h_t)$. During training, the agent may choose either the highest-probability action for better performance or sample probabilistically to ensure sufficient exploration. This forms the core \textbf{perception–action loop}: the agent obtains local observations, updates its GRU hidden state, and executes an action according to the softmax distribution. The environment then provides updated local observations (and informs whether a move was blocked), and the cycle repeats. As actions and observations alternate, position data accumulate within the network’s memory, enabling higher-level planning of direction and path.

\[
\text{Cycle per time step:}
\quad
\begin{cases}
\text{1. Acquire local observation } x_t \\
\text{2. Update hidden state } h_t \\
\text{3. Compute action distribution } \pi(a \mid h_t) \\
\text{4. Execute action } a_t \\
\text{5. Environment returns new observation } x_{t+1}
\end{cases}
\]

\begin{figure}[H]
    \centering
    \includegraphics[width=1.0\textwidth]{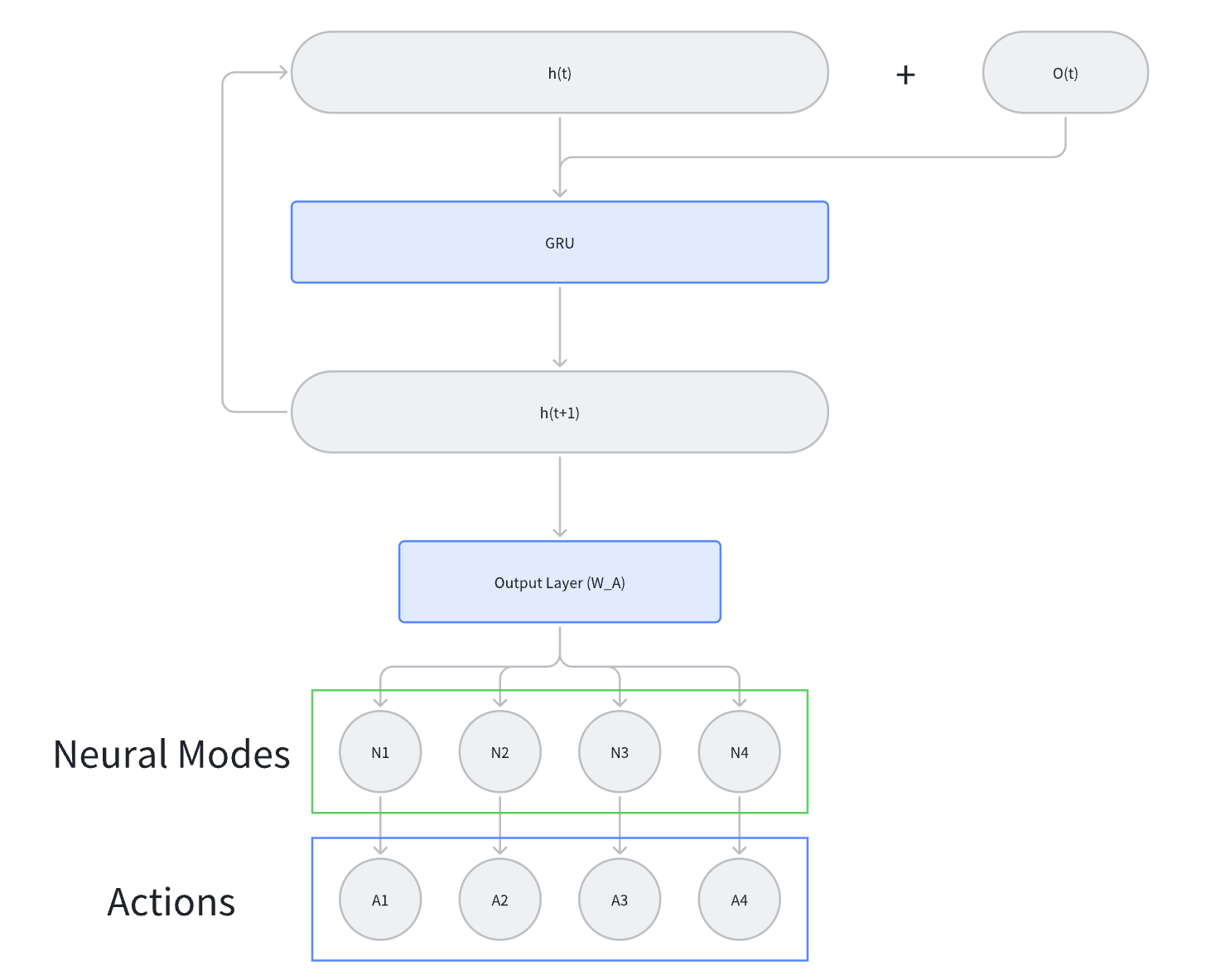}
    \caption{Gated Recurrent Unit–based architecture for memory and action selection.
 At each time step, the agent’s hidden state h(t) is concatenated with the local observation O(t) and passed into the GRU, yielding an updated hidden state h(t+1). A linear readout layer then maps h(t+1) to action logits for the four discrete moves, from which the agent chooses the action with the highest probability. This recurrent design ensures the agent retains and integrates information from past observations, supporting more informed navigation decisions under partial observability.}
\end{figure}

\subsubsection*{Training Procedure}

We adopt an \textbf{online policy optimization} strategy so that agents continuously refine their decision-making during exploration. Initially, with no maze knowledge, the agent explores all four actions randomly to gather diverse feedback. Once it reaches the goal for the first time, we \emph{do not} reset the GRU’s hidden state but only return its position to the maze start, allowing the agent to seek shorter paths or alternative strategies in the same maze. Over multiple round trips, this “Goal-Reset Mechanism” gradually refines discovered routes and also provides learning opportunities in unexplored areas. With each interaction, the network updates its gating parameters and output weights in response to success and failure, eventually forming more mature navigation policies.

To quantify performance, we measure the \textbf{path length of the final successful attempt}: if an agent continues to produce excessive detours even after several tries, it will receive a lower score. We also track success rates across different maze configurations. By combining these performance measures, we aim for the agent to maintain shorter paths and higher success rates in varied environments. We apply a central rank transformation to the path-length metric, giving higher weight to agents that excel, thereby guiding the evolutionary updates in the correct direction.

\begin{figure}[H]
    \centering
    \includegraphics[width=1.0\textwidth]{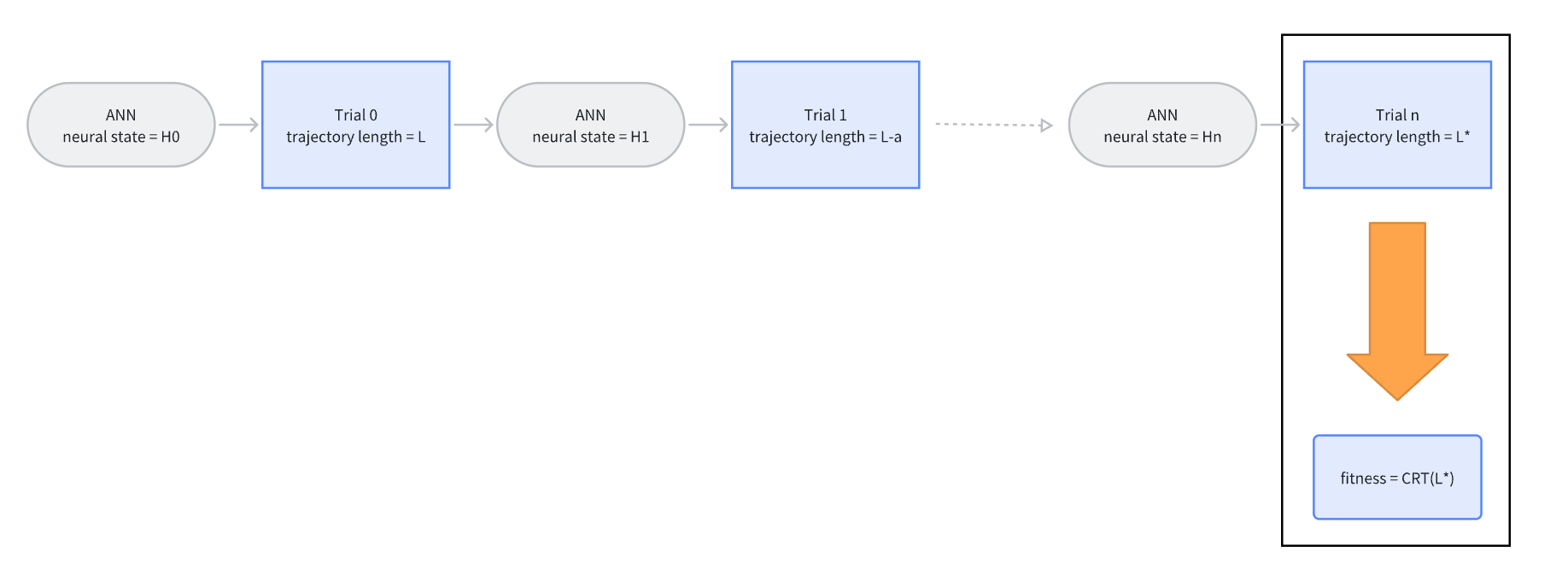}
    \caption{Fitness function based on the final successful path length and central rank transformation.
 During each agent’s finite lifespan, it executes multiple navigation trials within the same maze, carrying over its hidden state between trials. The final successful attempt yields a path length $L^*$. A central rank transformation is applied to $L^*$ across all candidates in the current generation, and the resulting rank serves as the agent’s fitness score in the Natural Evolution Strategy (NES) pipeline. Shorter final paths thus receive higher rankings, incentivizing more efficient navigation policies.}
\end{figure}

\subsubsection*{Evolutionary Algorithm \& Meta-RL}

We employ \textbf{Natural Evolution Strategy (NES)} to evolve the agent population: in each generation, we create tens to hundreds of thousands of mutated copies with perturbed parameters around a central solution. These copies interact in randomized mazes, and we then rank them based on metrics such as final path length. Top-ranked mutants update the population center solution for the next generation. Alongside NES, we integrate \textbf{meta-reinforcement learning (meta-RL)}: by repeatedly randomizing mazes and encouraging quick adaptation after initial exploration, agents gain global flexibility at the population level and local adaptability in individual mazes. In certain cases, if an agent recognizes key layout features in its first few steps, it can converge to an optimal or near-optimal route in subsequent attempts.

In summary, the \textbf{Agent Architecture \& Training} pipeline combines GRU-based memory for sequential decision-making under partial observability with NES+meta-RL joint optimization to evolve initial weights that generalize broadly. This approach allows agents to explore extensive collections of random mazes, gradually building robust internal representations of spatial factors like direction and distance. As shown in later sections (4.3 and 4.4), these learned representations exhibit signatures of “embodied cognition” in the form of stable dynamical structures and meaningful neural–behavior correspondences.

\begin{figure}[H]
    \centering
    \includegraphics[width=1.0\textwidth]{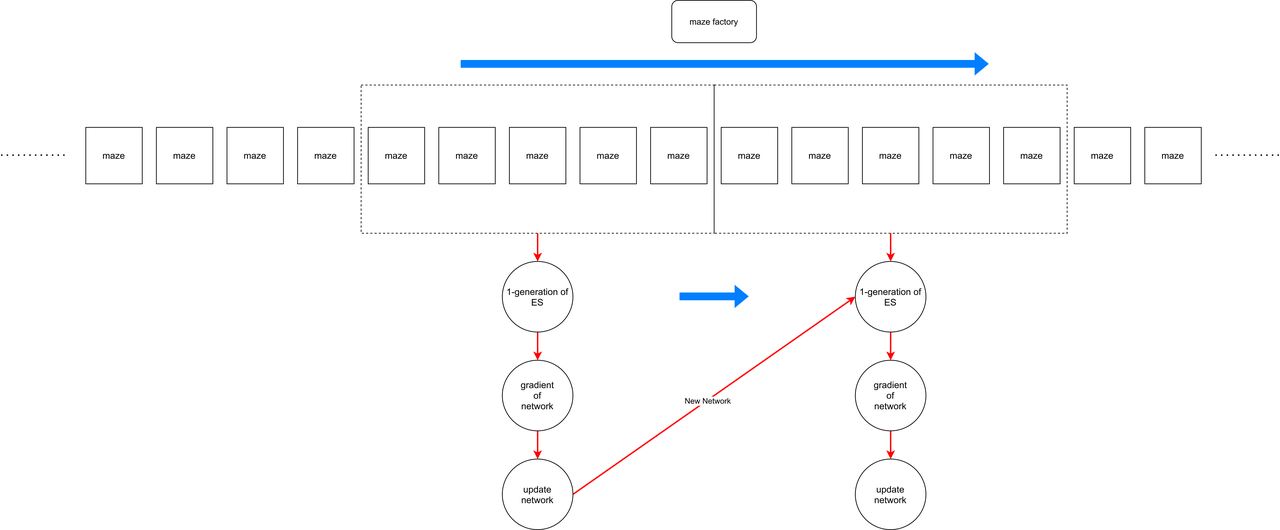}
    \caption{Natural Evolution Strategy (NES) for meta-training across randomly generated mazes.
 Each iteration begins with a “maze factory” producing 200 new environment seeds (dotted box). From the current central solution (red arrow), 100,000 mutated offspring are generated and evaluated on all 200 mazes. Each offspring’s fitness (based on navigation performance) contributes to a gradient update of the central solution’s parameters. Upon completion, all existing mazes are discarded, and the process repeats with fresh environments in the next iteration. This continual influx of novel mazes fosters robust generalization and avoids overfitting to a limited set of configurations.}
\end{figure}

\subsection{Hybrid Dynamical Systems Analysis}

Traditional research on neural network interpretability often treats the “input–network–output” pathway as unidirectional, focusing on how intermediate activations correlate with certain features. However, \emph{embodied intelligence} settings break this linearity: each agent action immediately alters future observations, creating \textbf{bidirectional coupling} between the network and environment—a “causal loop” in which the network both responds to and shapes the environment. Simply recording agent trajectories may not reveal whether the network genuinely drives behavior or merely reacts to environmental patterns. To address these causal complexities, we adopt the \textbf{Hybrid Dynamical Systems (HDS)} approach, integrating discrete environmental states with continuous neural states in a closed dynamical framework and tracking the agent’s time evolution holistically.

\begin{figure}[H]
    \centering
    \includegraphics[width=1.0\textwidth]{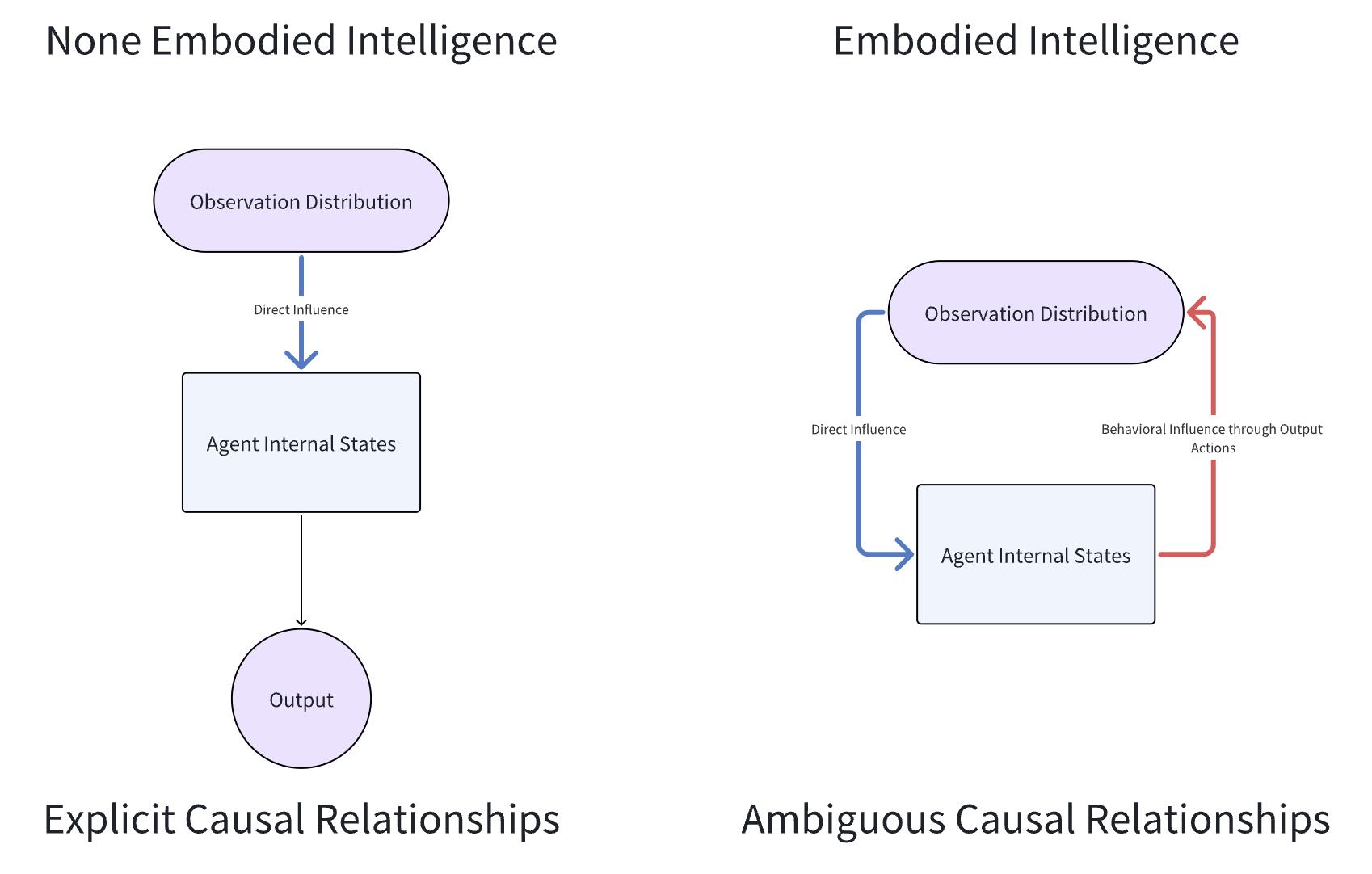}
    \caption{Contrasting causal structures in non-embodied versus embodied intelligence.
 (Left) In non-embodied systems, observations simply flow into the model’s internal states and generate outputs, resulting in relatively clear, unidirectional causal pathways.
 (Right) In embodied intelligence, the agent’s actions feed back into the environment, changing future observations. This bidirectional loop obscures cause-and-effect boundaries, leading to emergent macro-scale causal ambiguity in how information evolves.}
\end{figure}

\subsubsection{Overall Approach to solving the Causal Paradox}

In embodied navigation tasks, one can discretely represent the environment (e.g., maze coordinates $q_t$ and obstacle layouts) while treating the network’s hidden states $h_t \in \mathbb{R}^n$ as continuous. Analyzing how $q_t$ or $h_t$ evolves in isolation can lead to familiar explainable AI pitfalls—namely, seeing the network as merely responsive to external inputs, without acknowledging that network actions also steer future environmental states. The \textbf{HDS framework} directly addresses this issue:

\begin{enumerate}
    \item \textbf{State Definition}: Let $(q_t, h_t)$ be a point in the hybrid state space $Q \times X$, capturing both “agent location in the grid” and “network activations.”
    \item \textbf{Closed-Loop Evolution}: At each time step, the policy produces an action $a_t$, moving $q_t$ to $q_{t+1}$, while the network updates $h_t$ to $h_{t+1}$.
    \item \textbf{Unified Dynamical System}: The system thus evolves within $(Q \times X)$ in a fully closed loop, free of hidden external interventions. It can be examined as a single time-series process.
\end{enumerate}

\begin{figure}[H]
    \centering
    \includegraphics[width=1.0\textwidth]{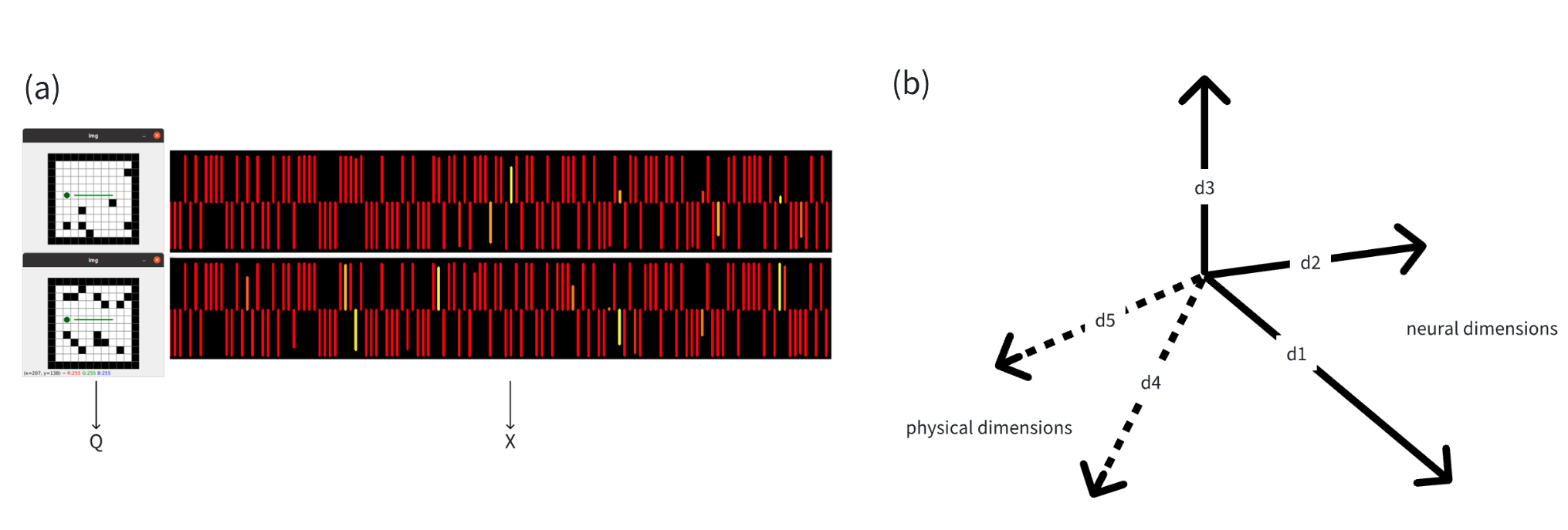}
    \caption{Rationale for hybrid dynamical sampling beyond direct environment trajectories.
(a) Illustration of raw maze trajectories (Q) alongside corresponding neural activations (X) for two different maze layouts. The maze panels (left) show agent movement (green circles and lines) in grids with obstacles (black squares); at each step, a locally observed state triggers updates in the network’s hidden units, depicted here as vertical bars (red/yellow) reflecting activation magnitudes. Directly recording such environment trajectories can introduce biases from suboptimal or exploratory paths, making it difficult to isolate the agent’s intended strategy.
(b) Conceptual diagram of the joint (Q×X) space, in which both physical coordinates (d4,d5) and neural dimensions (d1,d2,d3) jointly evolve in a closed loop. Examining this higher-dimensional hybrid system enables cleaner identification of stable or near-optimal strategy attractors, overcoming limitations of naive trajectory sampling.}
    \label{fig:my_figure}
\end{figure}

If the network has truly mastered an optimal strategy, we might see stable \textbf{attractors} in $(Q \times X)$—for example, a \textbf{limit cycle}, indicating that the joint agent–environment trajectory repeatedly cycles through the same states. This observation goes beyond noting, “the agent repeatedly traversed a path,” because synchronized periodicity in both neural and physical states rules out mere transient memory or passive environment constraints. An attractor also resists small perturbations, indicating a deeply internalized dynamical solution.

\subsubsection{System Modeling and State Evolution}

We denote each time step as $(q_t, h_t) \in Q \times X$. Specifically:

\begin{itemize}
    \item \textbf{$Q$}: Discrete maze coordinates (including obstacles, goals, etc.). If the agent tries to step into a wall or outside the grid, it remains in place.
    \item \textbf{$X$}: Neural hidden states $h_t \in \mathbb{R}^n$, updated via gating mechanisms based on past activations and current observations.
\end{itemize}

If we let $f(\cdot)$ represent the network update, $g(\cdot)$ map network states to actions, and $h(\cdot)$ define the environment’s response to actions, the time-step equations become:

\[
\begin{cases}
\mathbf{h}_{t+1} = f(\mathbf{h}_t, \mathrm{obs}(q_t)), \\
a_t = g(\mathbf{h}_t), \\
q_{t+1} = h(q_t, a_t).
\end{cases}
\]

This defines a complete closed dynamical system from $(q_t, h_t)$ to $(q_{t+1}, h_{t+1})$. Starting from any initial state $(q_0, h_0)$, the system traces out a trajectory. If a specific state reappears, forming a cycle, it signifies a \textbf{limit cycle} in hybrid space, which often corresponds to an optimal strategy attractor.

\subsubsection{Limit Cycle Detection and Lyapunov Stability Analysis}

\paragraph{(1) Basic Detection}

We first run fully trained agents multiple times in the same or similar mazes (using “Goal-Reset” to preserve hidden states but reset physical location) and record $(q_t, h_t)$. We look for a period $T$ such that

\[
\|(q_{t+T}, \mathbf{h}_{t+T}) - (q_t, \mathbf{h}_t)\| \,\le\, \epsilon
\]

holds repeatedly. Techniques like sliding-window matching or similarity metrics help find repeating subsequences, with $\epsilon$ controlling the identity threshold. If periodicity is detected, the system has a closed-loop cycle in hybrid space. When this loop represents an \textbf{optimal action cycle}, the agent typically revisits nearly the same physical coordinates and neural activations over multiple rounds.

\paragraph{(2) Small Perturbations and Lyapunov Exponents}

Finding a loop does not prove it is attractively stable. To investigate, we introduce small perturbations $\delta$ at selected cycle points and observe if deviations shrink or grow:

\begin{itemize}
    \item If deviations decay exponentially, local Lyapunov exponents are negative, indicating an attractor.
    \item If they spread or oscillate randomly, the loop may be unstable or chaotic.
\end{itemize}

We estimate Lyapunov exponents by examining the logarithmic divergence of trajectories around the loop for many sampled points. Mature, stable cycles often yield negative exponents (e.g., –0.1 to –0.4), while exponents near zero or positive suggest instability.

\begin{figure}[H]
    \centering
    \includegraphics[width=1.0\textwidth]{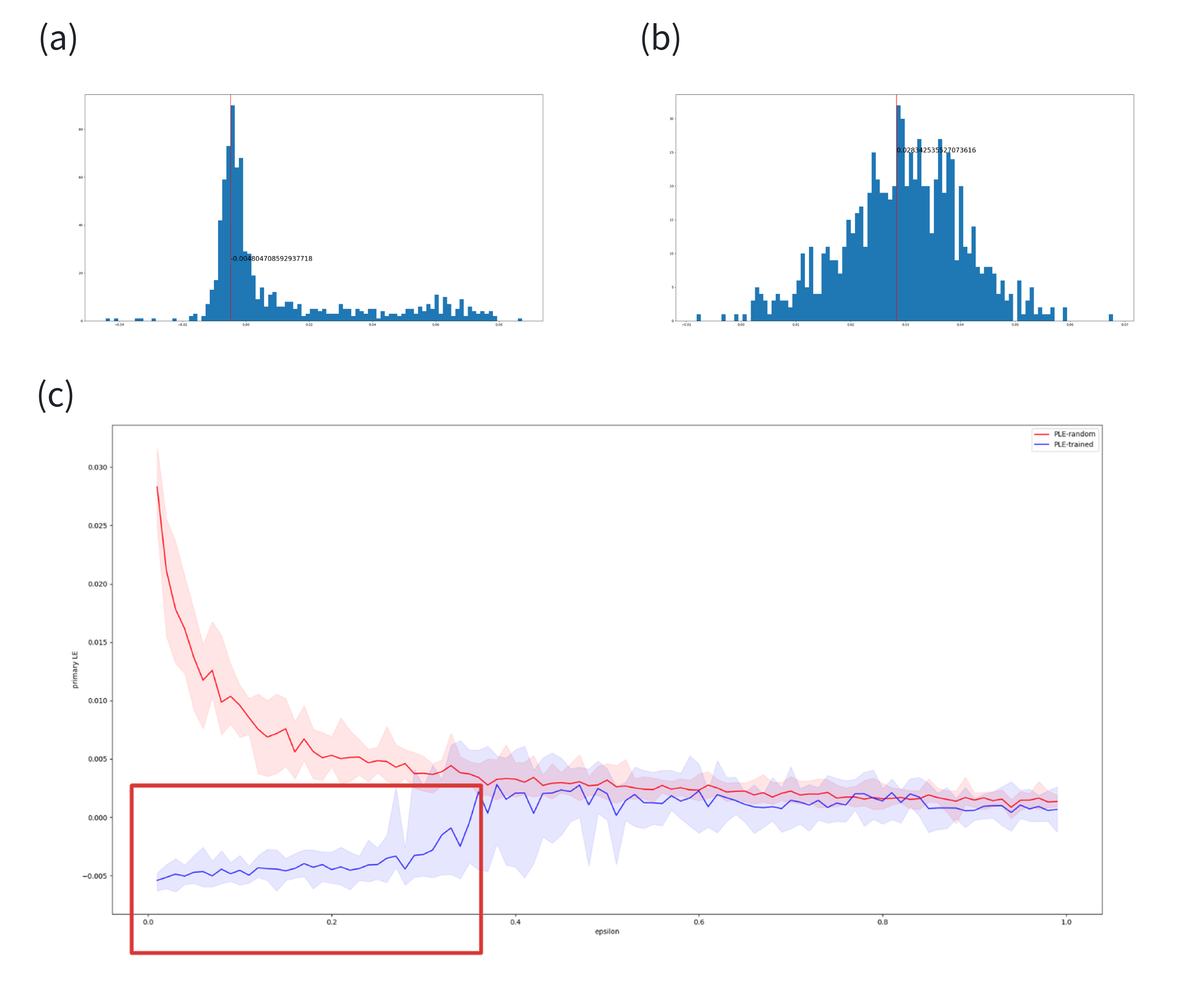}
    \caption{Lyapunov exponent (LE) distributions and perturbation-based stability analysis.
(a) Histogram of LEs for the trained GRU when the perturbation magnitude is epsilon=0.01; the peak LE is around -0.004, indicating locally stable or attracting dynamics.
(b) In contrast, a random GRU under the same small perturbation exhibits a peak LE of approximately +0.028, suggesting locally expanding or less stable trajectories.
(c) As perturbation magnitude increases (horizontal axis), the LE for the trained system (blue) remains negative over a wider range-highlighted by the red box-whereas the random system’s LE (red) crosses to positive more readily. This difference delineates the “invariant set” region in which the trained network maintains stable limit cycles or attractors, while the random network lacks such robust stability.}
\end{figure}

\subsubsection{Cyclic Stimulation Verification}

To further rule out random environmental changes or noise, we introduce \textbf{cyclic stimulation}, which tests whether a loop is genuinely an attractor by removing external variability:

\begin{enumerate}
    \item \textbf{Extract Key Input Sequences}

    Gather representative local observations (e.g., visual streams) from the agent’s optimal or near-optimal paths, forming a fixed input sequence.

    \item \textbf{Block Real Observations}

    Feed only this artificial sequence to the network, bypassing true maze input.

    \item \textbf{Reset Initial States}

    Initialize $(q_0, h_0)$ to chosen or random conditions, then feed the fixed input repeatedly. If a true attractor exists, the system should return to its periodic trajectory.

    \item \textbf{Test Attraction Range}

    If the same loop reemerges under various starting states and minor noise, it has a broad attraction domain; if only specific states work, the attractor is narrower or less robust.
\end{enumerate}

\begin{figure}[H]
    \centering
    \includegraphics[width=1.0\textwidth]{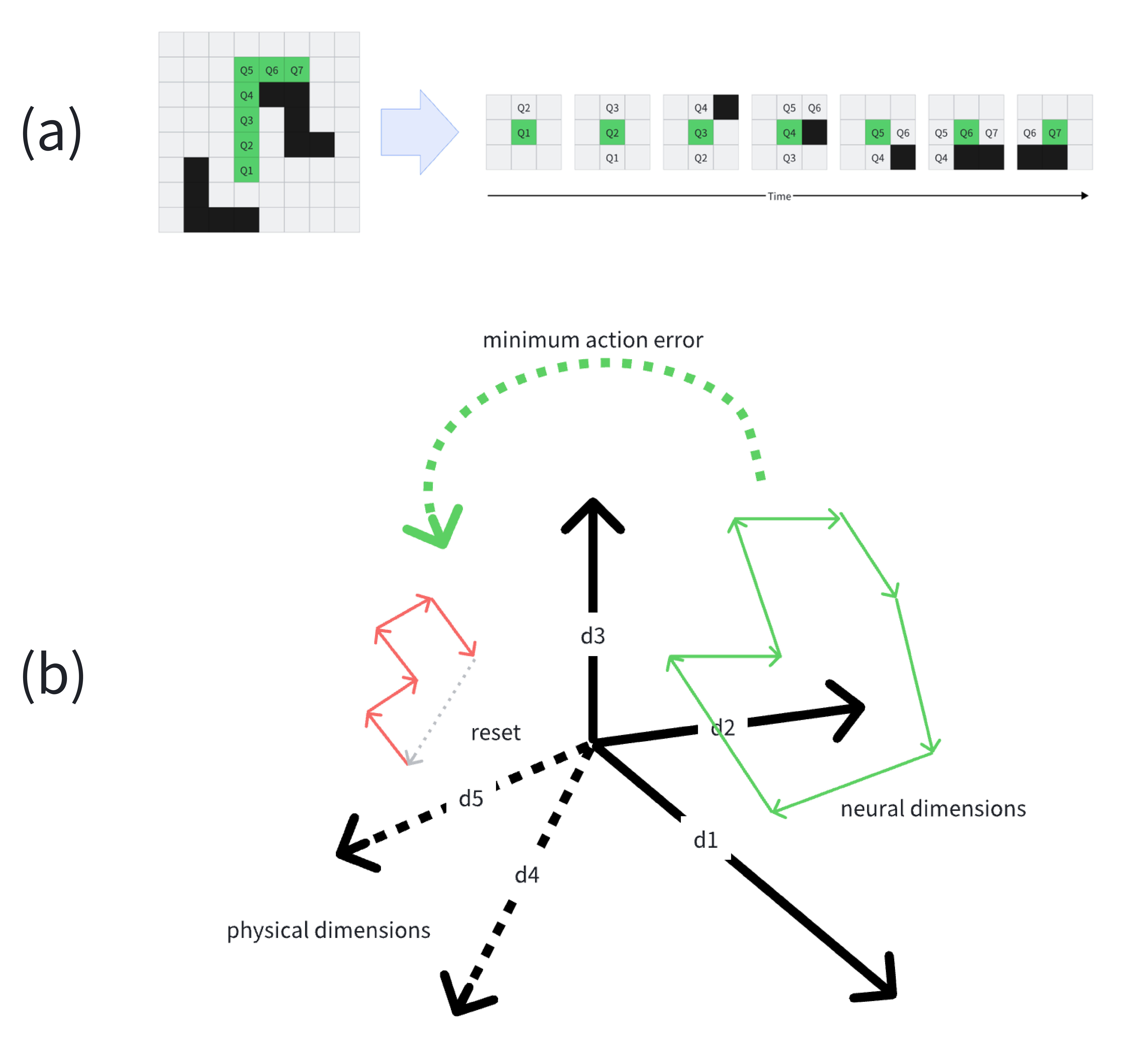}
    \caption{Cyclic stimulation approach for verifying hybrid limit cycles in Q×X space.
(a) During cyclic stimulation, the same input sequence (e.g., observed maze cells at each time step) is repeatedly fed to the agent from multiple random initial states. The agent transitions through cells Q1 to Q7, but only trajectories whose action outputs match the ground-truth sequence are retained.
(b) Conceptual view of the resulting trajectories in hybrid space, where physical coordinates (d4,d5, dashed axes) and neural dimensions (d1,d2,d3, solid axes) jointly evolve. Each random initialization may produce a slightly different neural path (green or red), but only those with “minimal action error” converge to the same limit cycle as the real-world behavior. This method confirms that the agent’s stable attractor in the RNN state space underpins its observed optimal navigation loop.}
\end{figure}

Because it excludes changing environment factors, cyclic stimulation helps confirm that a periodically repeating loop is a real “internalized strategy” rather than an incidental result of environment–network interplay. If a loop cannot be reactivated with a fixed sequence or is easily disrupted by small perturbations, it likely is not a stable attractor. Conversely, if cyclic stimulation repeatedly triggers the same cycle and the corresponding Lyapunov analysis confirms negative exponents, then the agent has integrated a deeply stable strategy.

\subsubsection{Summary and Connection to Results}

By combining \textbf{HDS modeling, limit cycle detection, Lyapunov stability metrics, and cyclic stimulation}, we can both visualize and rigorously validate emergent optimal strategy cycles in hybrid space. Unlike purely environmental sampling that may capture suboptimal or incomplete routes, HDS allows \emph{active} probing of the agent’s genuine, internalized strategies. In our \textbf{Results} section, we use these HDS-sampled neural states alongside \textbf{Ridge Representation} to quantify neural–behavioral alignment, and we show through CCA and intervention experiments how certain neural dimensions robustly encode spatial information. If intervening in those dimensions disrupts behavior, it confirms that the agent has deeply embedded spatial concepts within its neural architecture—what we term a “deep embodied understanding.”

\subsection{Behavioral \& Neural Representation Methods}

To analyze how agent behavior relates to neural representations, we first require a unified, robust way to encode navigation paths. Agents may produce trajectories of varying lengths and shapes, and directly using coordinate sequences can complicate metrics and comparisons. We address this by using \textbf{Ridge Representation}, which embeds each two-dimensional path in a fixed-resolution grayscale image, preserving key geometric features and enabling distance-based comparisons.

\subsubsection*{Ridge Representation of Paths}

Consider a navigation trajectory as a series of discrete points $\{p_1, p_2, \ldots, p_L\}$. We define a fixed-size image grid (e.g., 21$\times$21 pixels) and, for each trajectory point $p_i$, generate a linearly attenuating “radiation field.” Specifically, if the Euclidean distance from a pixel $(u,v)$ to $p_i$ is $d_i(u,v)$, the intensity at that pixel is $\max\{0, \alpha - \beta \cdot d_i(u,v)\}$, where $\alpha,\beta$ control the attenuation. Once a pixel’s distance exceeds a threshold, its value becomes zero. We then adopt a \textbf{maximum-value fusion} rule to aggregate the influence of all $p_i$ onto each pixel, resulting in a clear “ridge” representing the trajectory’s position and direction.

Consequently, each path is transformed into a grayscale image of uniform size. Even paths with very different lengths or shapes receive comparable, fixed-dimensional representations. Minor local translations or deformations cause moderate image differences, allowing metrics defined in this space to better capture geometric path relationships. We use the \textbf{L2 norm} (Euclidean distance) to quantify differences between two grayscale images: similar paths yield small pixel-level discrepancies, while highly distinct paths yield larger ones. Because Euclidean distance satisfies standard metric axioms (non-negativity, symmetry, triangle inequality), paths can be compared within a proper metric space.

Implementation details must address normalization or truncation of grayscale values to avoid saturation and carefully set the attenuation parameters $(\alpha,\beta)$. Excessively large $(\alpha,\beta)$ can make the entire image overly bright and blur path distinctions; too small, and minor variations may appear disproportionately large. In practice, \textbf{Ridge Representation} effectively preserves overall direction, local turns, and other geometry while unifying different trajectories into a consistent image-based format—facilitating dimensionality reduction (e.g., PCA) and neural correlation analyses (e.g., CCA).

\subsubsection*{Relating Trajectories and Neural States}

In preliminary evaluations, we performed \textbf{Principal Component Analysis (PCA)} on both the neural states (e.g., GRU hidden vectors) and the corresponding Ridge images of each trajectory. Visual inspection in the first few principal components revealed ring-like or stratified distributions in both spaces, suggesting they might share some geometric structure in lower dimensions. However, such visual similarities alone do not constitute definitive proof of a deeper alignment.

To examine this systematically, we introduced \textbf{Canonical Correlation Analysis (CCA)}, which identifies linear projection directions maximizing correlation between two datasets—here, neural states and Ridge-based behavioral features. If the agent has genuinely encoded spatial elements such as direction and distance, the first several pairs of canonical projections should exhibit notably high correlation (e.g., above 0.8), not just in a limited 2D or 3D visualization. Conversely, if high correlation arises in only a few dimensions or hovers near noise levels, it would indicate minimal structural coupling.

\begin{figure}[H]
    \centering
    \includegraphics[width=1.0\textwidth]{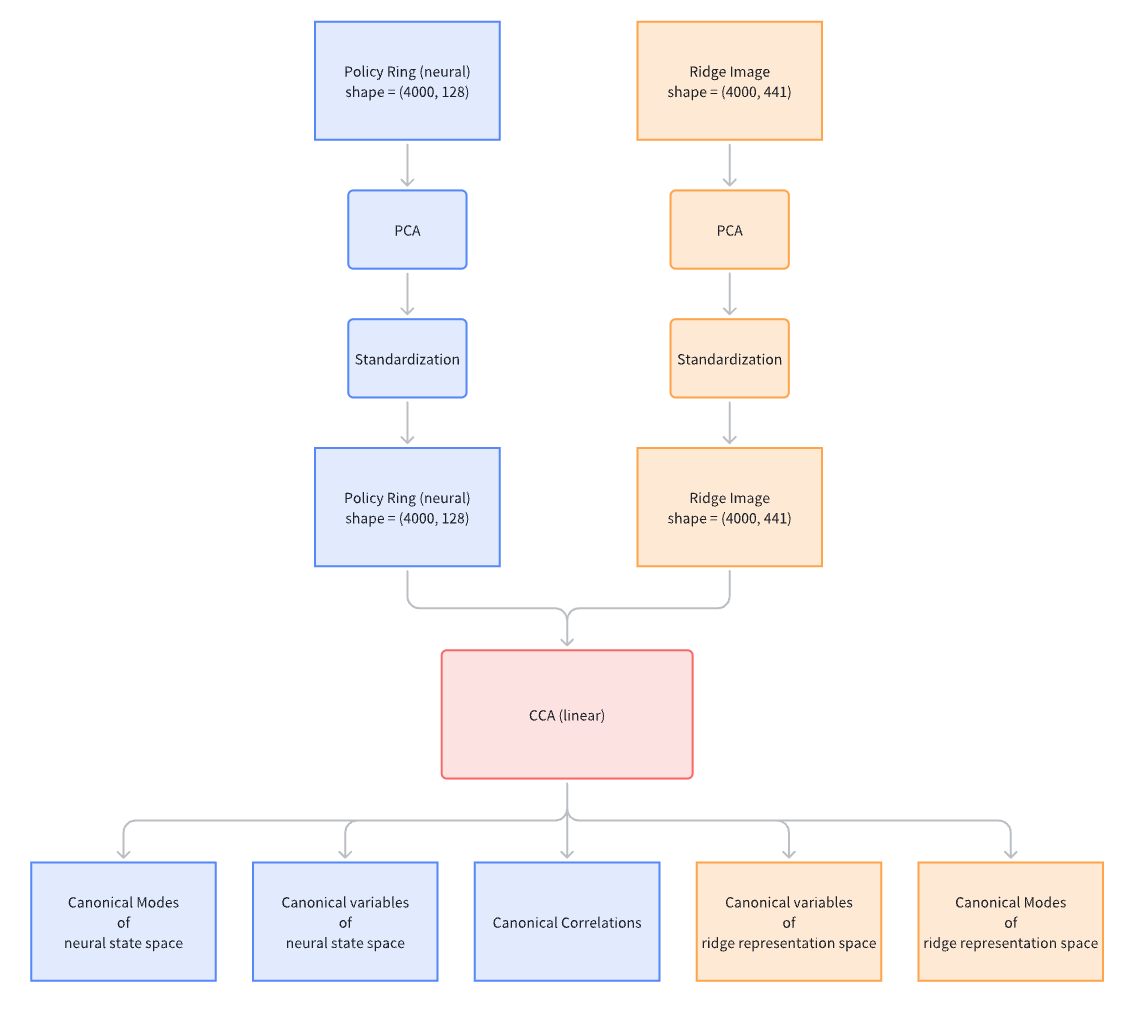}
    \caption{Workflow for comparing neural states and Ridge-based representations via CCA.
 Both the agent’s neural states (left) and the corresponding Ridge images (right) undergo Principal Component Analysis (PCA) followed by standardization to reduce dimensionality and eliminate scale differences. The resulting embeddings (e.g., 4,000 samples × 128 neural features, and 4,000 samples × 441 Ridge pixels) are then passed into a linear Canonical Correlation Analysis (CCA). This procedure finds pairs of canonical modes in each modality that maximize mutual correlation, revealing high-dimensional correspondences between the agent’s internal “policy ring” and its explicit behavioral path representations.}
\end{figure}

\subsubsection*{Counterfactual \& Intervention Experiments}

Even if strong neural–behavior correlations emerge, we still need to test \emph{causality}: Do these correlated neural dimensions genuinely drive navigation decisions? To investigate, we use \textbf{Intervention} and \textbf{Counterfactual} approaches, altering the identified “critical” neural dimensions to see if performance degrades.

\begin{figure}[H]
    \centering
    \includegraphics[width=1.0\textwidth]{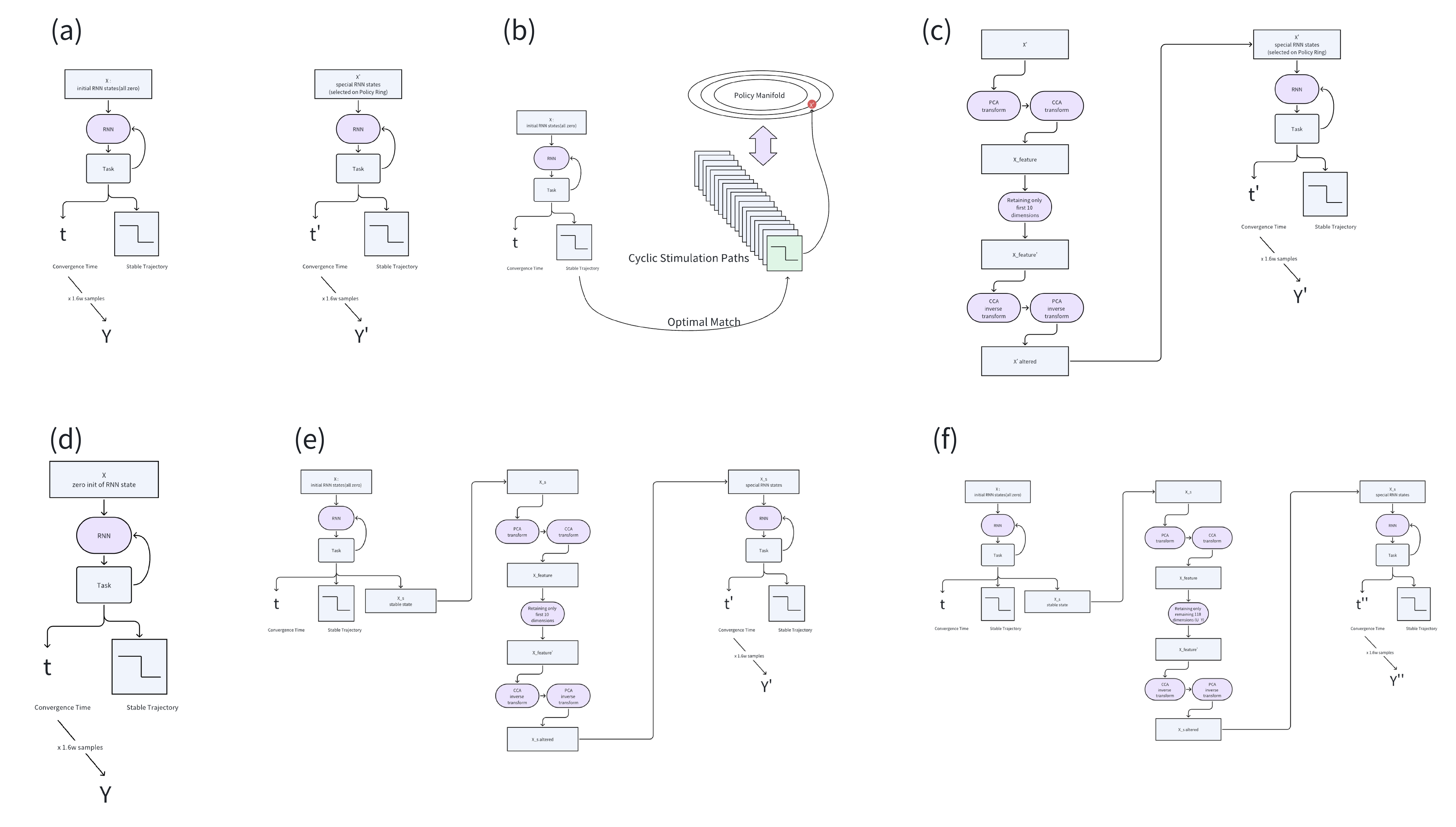}
    \caption{Schematic of various counterfactual and intervention procedures used to verify causal significance of neural dimensions.
(a) Overview of the core logic: each experiment measures how modifying or injecting particular hidden states (X) influences the agent’s convergence time and resulting trajectory (t,Y)(t, Y).
(b) Procedure for identifying an “optimal” hidden-state vector X' from the policy manifold (e.g., via cyclic stimulation). This vector is then inserted into the agent’s RNN to see if it improves navigation efficiency.
(c) Feature-space manipulation: a PCA/CCA transform is applied to X', and only certain high-correlation dimensions are retained or altered before re-projecting them back into the RNN. The agent’s performance under this partial intervention is compared with baseline.
(d) Reference experiment zeroing out the RNN’s hidden state. The resulting performance serves as a control for interventions that remove key learned signals.
(e) First counterfactual scenario: injecting a precomputed neural state X' while randomizing non-critical dimensions, examining whether the agent retains stable trajectories and fast convergence.
(f) Second counterfactual scenario: randomizing the critical dimensions in X' and preserving others, testing whether performance reverts toward baseline convergence times (slower or less stable), thereby confirming the causal role of the targeted dimensions.}
\end{figure}

\begin{enumerate}
    \item \textbf{Dimension Selection}

    We pick the neural components most strongly correlated via CCA (e.g., those surpassing a correlation threshold of 0.8). Alternatively, we select dimensions showing high sensitivity or variance, dividing them into “critical” and “non-critical” groups.

    \item \textbf{Intervention Methods}

    \begin{itemize}
        \item \emph{Zeroing} all values in the chosen dimension(s).
        \item \emph{Randomizing} them (replacing original values with noise).
        \item \emph{Shifting or scaling} them, or even copying values from other dimensions.
    \end{itemize}
    These manipulations directly target the most correlated neural axes while preserving the rest.

    \item \textbf{Performance Measurement}

    We re-run navigation tasks (in the same or similar mazes) and record path length, success rate, and other metrics. If performance markedly declines, those neural dimensions have causal significance.

    \item \textbf{Control \& Timing}

    We sometimes apply the same intervention to low-correlation dimensions or random networks to confirm they do not exhibit comparable performance drops. We also vary when and how interventions are applied (e.g., continuously, intermittently), clarifying which neural signals are crucial for decision-making.
\end{enumerate}

In essence, \textbf{Counterfactual \& Intervention} experiments elevate strongly correlated neural dimensions from “statistical associations” to potential \textbf{causal drivers}. If manipulating a given dimension directly impairs the agent’s navigation strategy, that dimension likely encodes part of the agent’s “embodied world model.” This methodology not only reinforces conclusions from neural–behavior correlation but also advances our understanding of model interpretability and controllability.

\section{Additional Information}

\textbf{Acknowledgements}: We would like to thank colleagues from the Tsinghua Laboratory of Brain and Intelligence for their valuable discussions and feedback on earlier versions of this manuscript.

\textbf{Author Contributions}: Lijin designed the experiment, implemented the models, and conducted the data analysis, and wrote the manuscript. Liujia contributed to the theoretical framework, supervised the research.

\textbf{Competing Interests}: The authors declare no competing interests.

\end{document}